%% file: main.tex
\documentclass[manuscript]{acmart}
\AtBeginDocument{%
  \providecommand\BibTeX{{%
    \normalfont B\kern-0.5em{\scshape i\kern-0.25em b}\kern-0.8em\TeX}}}

\setcopyright{acmcopyright}
\copyrightyear{2018}
\acmYear{2018}
\acmDOI{XXXXXXX.XXXXXXX}

\acmConference[Conference acronym 'XX]{Make sure to enter the correct
  conference title from your rights confirmation emai}{June 03--05,
  2018}{Woodstock, NY}
\acmBooktitle{Woodstock '18: ACM Symposium on Neural Gaze Detection,
 June 03--05, 2018, Woodstock, NY} 
\acmPrice{15.00}
\acmISBN{978-1-4503-XXXX-X/18/06}

\usepackage{graphicx}
\usepackage{float}
\usepackage{graphicx,caption}
\usepackage{multirow}

\usepackage{colortbl} 
\usepackage{soul}
\sethlcolor{green}
\usepackage{wrapfig}

\newcommand\wordcount{
    \immediate\write18{texcount -sub=section \jobname.tex  | grep "Section" | sed -e 's/+.*//' | sed -n \thesection p > 'count.txt'}
(\input{count.txt}words)}

\NewDocumentCommand{\revanth}
{ mO{} }{\textcolor{brown}{\textsuperscript{\textit{Revanth}}\textsf{\textbf{\small[#1]}}}}

\NewDocumentCommand{\daniel}
{ mO{} }{\textcolor{green}{\textsuperscript{\textit{Daniel}}\textsf{\textbf{\small[#1]}}}}

\newcommand{\name}{\textsc{SmartBook}}

\NewDocumentCommand{\heng}
{ mO{} }{\textcolor{red}{\textsuperscript{\textit{Heng}}\textsf{\textbf{\small[#1]}}}}

\renewcommand\footnotetextcopyrightpermission[1]{} 

\begin{document}
\title{\name{}: AI-Assisted Situation Report Generation for Intelligence Analysts} 
\pagestyle{plain}
\author{Revanth Gangi Reddy}
\email{revanth3@illinois.edu}
\affiliation{%
  \institution{University of Illinois at Urbana-Champaign}
  \city{Champaign}
  \state{Illinois}
  \country{USA}
}

\author{Daniel Lee}
\affiliation{%
  \institution{University of Calgary}
  \city{Calgary}
  \city{Alberta}
  \country{Canada}}
\email{daniel.lee1@ucalgary.ca}

\author{Yi Fung}
\email{yifung2@illinois.edu}
\affiliation{%
  \institution{University of Illinois at Urbana-Champaign}
  \city{Champaign}
  \state{Illinois}
  \country{USA}
}

\author{Khanh Duy Nguyen
 }
\email{knguye71@illinois.edu}
\affiliation{%
  \institution{University of Illinois at Urbana-Champaign}
  \city{Champaign}
  \state{Illinois}
  \country{USA}
}

\author{Qi Zeng}
\email{qizeng2@illinois.edu}
\affiliation{%
  \institution{University of Illinois at Urbana-Champaign}
  \city{Champaign}
  \state{Illinois}
  \country{USA}
}

\author{Manling Li}
\email{manling.li@northwestern.edu}
\affiliation{%
  \institution{Northwestern University}
  \city{Chicago}
  \state{Illinois}
  \country{USA}}

\author{Ziqi Wang}
\email{ziqiw9@illinois.edu}
\affiliation{%
  \institution{University of Illinois at Urbana-Champaign}
  \city{Champaign}
  \state{Illinois}
  \country{USA}
}


\author{Clare Voss}
\affiliation{%
  \institution{Army Research Laboratory}
  \city{Adelphi}
  \state{Maryland}
  \country{USA}}
\email{clare.r.voss.civ@army.mil}

\author{Heng Ji}
\affiliation{%
  \institution{University of Illinois at Urbana-Champaign}
  \city{Champaign}
  \state{Illinois}
  \country{USA}
}

\email{hengji@illinois.edu}

\input{nature_sections/abstract}
\maketitle

\input{nature_sections/introduction}

\input{nature_sections/methodology}

\input{nature_sections/results}

\input{nature_sections/discussion}

\input{nature_sections/supplementary}

\bibliographystyle{ACM-Reference-Format}
\bibliography{sample-base}



\end{document}

%% file: nature_sections/abstract.tex
\begin{abstract}
Timely and comprehensive understanding of emerging events is crucial for effective decision-making; automating situation report generation can significantly reduce the time, effort, and cost for intelligence analysts. In this work, we identify intelligence analysts' practices and preferences for AI assistance in situation report generation to guide the design strategies for an effective, trust-building interface that aligns with their thought processes and needs. Next, we introduce \name{}, an automated framework designed to generate situation reports from large volumes of news data, creating structured reports by automatically discovering event-related strategic questions. These reports include multiple hypotheses (claims), summarized and grounded to sources with factual evidence, to promote in-depth situation understanding.
Our comprehensive evaluation of \name{}, encompassing a user study alongside a content review with an editing study, reveals \name{}'s effectiveness in generating accurate and relevant situation reports. Qualitative evaluations indicate over 80\% of questions probe for strategic information, and over 90\% of summaries produce tactically useful content, being consistently favored over summaries from a large language model integrated with web search. The editing study reveals that minimal information is removed from the generated text (under 2.5\%), suggesting that \name{} provides analysts with a valuable foundation for situation reports.

\end{abstract}

%% file: nature_sections/introduction.tex
\section{Introduction}\label{sec1}

In today's rapidly changing world, intelligence analysts face the constant challenge of staying informed amidst an overwhelming influx of news, rumors, and evolving narratives. To understand unfolding events, it is essential to extract common truths from heterogeneous data sources. Currently, intelligence analysts prepare \textit{situation reports} that provide an overview of the state of affairs, potential risks or threats, and perspectives, along with recommended actions to guide action planning and strategic development~\cite{world2020coronavirus,world2022emergency}. Situation reports are expected to present salient information about key events and trends with a clear logical structure to facilitate understanding, tracking, and easy access to information.
 The downstream consumers of these reports (general public, decision-makers, and stakeholders) are not just looking for raw information; they are seeking clarity, context, and actionable insights. However, these reports, being manually crafted, come with limitations: they are time-consuming to produce~\cite{national2019decadal}, might exhibit biases~\cite{biases1, belton2020cognitive}, factual errors~\cite{trustllm2024}, and may be restricted due to the large volume of information they must process~\cite{doggette2020information, overload1}.
Intelligence analysts spend a lot of time sifting through vast and complex data sources~\cite{golson2018resist, national2011intelligence} ({\em e.g.,} social media rumor propagation, news outbreak, background statistics, domain expert opinions, etc.), and readers often grapple with information that's either too dense or not sufficiently comprehensive. Given the sheer volume of information, it is challenging for analysts to do deep analysis and critical thinking, to be able to formulate strategic questions and generate insights.
We hypothesize that automatic situation report generation can bridge the gap between reading and writing intelligence information. 
However, existing summarization approaches~\cite{fabbri-etal-2019-multi, abdi2017query, li2014improving} generate plain-form text and do not yield the details, structures, and high-level strategic information to support decision-making for the downstream consumer. In particular, current large language model-based approaches such as GPT-4/ChatGPT~\cite{openai2021gpt35} may return no answers or incorrect answers to user questions as they do not update information in real-time and lack fact-checking mechanisms ({e.g.,} Table~\ref{tab:chatgpt}). 

To establish a basis for the automated generation of situation reports, we carry out a formative study (see \S{\ref{sec:formative_study}}) to grasp intelligence analysts' expectations from AI-driven systems. Our findings reveal that analysts are open to AI assistance, seek clarity in AI operations, and vary in their desire to control the AI system. 
To comprehend the composition process of situation reports, we augmented the initial study's findings with collaborative design sessions (see \S{\ref{sec:collaborative_design}}), aiming to define practical design strategies and recommendations, particularly on how human intelligence analysts navigate, research, and write their reports. On a high level, we infer that the system interface must visually align with analysts' sequential thought processes, provide transparency of information flow to build trust, and enable analysts to have some level of control over the AI tool's processes. These conclusions consequently guided the design strategies for automating the generation of situation reports.

\input{tables/chatgpt}

\begin{figure}[t]
\includegraphics[width=\textwidth]{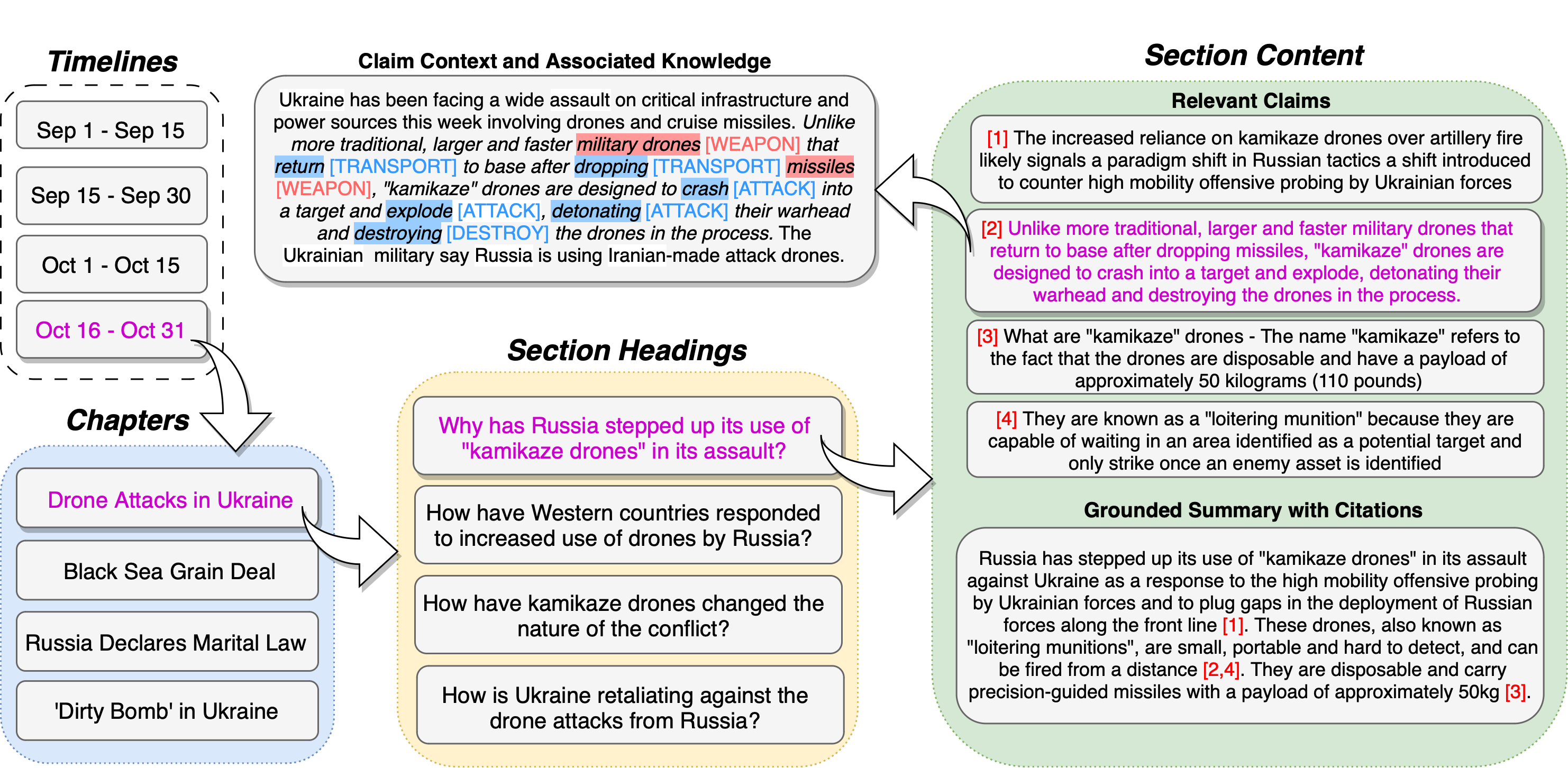}
\caption{Figure showing an example from SmartBoook for the Ukraine-Russia Crisis. \name{} is organized on a timeline, with 2-week time spans containing chapters and corresponding sections. The section headings are strategic questions and each section content consists of a grounded summary with links to relevant claims, each highlighted for factual evidence and knowledge elements. The italicized entities and events, which comprise the associated knowledge elements, are highlighted in red and blue respectively.} 
\label{fig:1}
\end{figure}

Building on the foundation set by the formative study and collaborative design, we present \textbf{\name{}}, a framework designed to assist human analysts who author situation reports. SmartBook ingests data from multiple sources to generate a comprehensive report with information updated regularly. Human analysts typically source information by asking questions based on their own understanding of the situation. However, these questions can be static or too vague, and become outdated as the situation evolves rapidly. In contrast, \name{} automatically discovers questions 
important for situation analysis and gathers salient information for generating the report. For all questions about a major event, the report contains summaries with tactical information coming from relevant claims, presented with local context and links to source news articles. \name{} structures data in a manner that mirrors current intelligence analyst workflows--breaking down events into timelines, chapters, and question-based sections. Each section contains a grounded query-focused summary with its relevant claims. This intuitive structure facilitates easier assimilation of information for both reading and writing. Designed with a human-centered approach, our goal is to augment the capabilities of human analysts rather than replace them. Figure \ref{fig:1} shows an example from \name{} for the Ukraine-Russia crisis, with the structured hierarchy of timespans, chapters, and corresponding sections.

In our comprehensive evaluation of \name{}, we conducted two complementary studies: a utility study (in \S{\ref{sec:utility_eval}}) for assessing usability and interaction, and a content review (in \S{\ref{sec:content_review_study}}) for examining the quality of text summaries. The user study, involving intelligence analysts and decision-makers, focused on key research questions to explore \name{}'s usability, intuitiveness, and effectiveness in situation report generation. Participants engaged in semi-structured interviews and post-study questionnaires, reflecting on their experience with the system. The content review complemented this by assessing the readability, coherence, and relevance of \name{}-generated reports, including an editing study where an expert analyst revised the summaries to meet their standards of acceptability.  The primary goal of this process was to evaluate the viability of using \name{} as a tool for creating preliminary drafts of situation reports. The findings indicate that the content generated by \name{} was mostly accurate, requiring minimal edits to correct a few factual errors. However, most of the effort in the editing process was focused on adding evidence to support the summaries. The results of the editing study suggest that while \name{} provides a solid foundation, it significantly benefits from human refinement. 
The contributions of this work are as follows:
\begin{itemize}
    \item A comprehensive formative study and collaborative design process for identifying the design strategies to guide the automated generation of situation reports.
    \item \textbf{\name{}}, an automated framework that generates comprehensive, up-to-date situation reports from various sources and presents them in an intuitive and user-friendly manner. It identifies critical strategic questions, ensuring that downstream readers receive targeted, relevant, and evidence-grounded information to aid their decision-making processes.
    \item A thorough utility evaluation involving intelligence analysts and  decision-makers investigating the usability of the system.
    \item A content review to grade the quality of the information generated, along with an editing study to understand how viable \name{}  is for producing preliminary first drafts of situation reports.    
\end{itemize}
    

%% file: tables/chatgpt.tex
\begin{table}[t]
\centering
\renewcommand{\arraystretch}{1.5}
\small
\begin{tabular}{p{52em}}
\hline
\textbf{Question}: 
Is Russian positioning forces 
for possible follow-on actions into Ukraine after the initial invasion? \\
\hline
\textbf{GPT-4}: As an AI language model, I cannot provide real-time information on current events. However, it is essential for intelligence analysts to monitor the activities of various countries, including troop movements and military exercises,to assess potential threats and anticipate future actions. To determine if Russia is positioning forces for possible actions into Ukraine, one must analyze the latest intelligence reports, satellite imagery, and open-source information. Additionally, understanding the political and historical context of the region can provide valuable insight into the intentions of the countries involved.\\
\hline
\textbf{SmartBook}: The arrival of Russian forces in Slavutych came after several days of shelling against the city, which is strategically located close to the Dnieper River, about 95 miles (150 kilometers) north of Kyiv and close to the border with Belarus.
Russia has been using Belarus as a springboard for many of its air operations in Ukraine, according to intelligence collected by NATO surveillance planes flying over the Polish-Ukrainian border and radar seen by CNN.
So far, however, US officials have not seen Belarusian troops ``being readied to move into Ukraine" or ``that they are moving or are in Ukraine", a senior US defense official said Monday, adding that the forces inside Ukraine are Russian.\\
\hline
\end{tabular}
\caption{Example of situation understanding for the Ukraine-Russia crisis. GPT-4 is incapable of generating situation reports because it has limited knowledge of up-to-date events, while Smartbook is able to generate timely, multi-source, and trustworthy answers.
}
\label{tab:chatgpt}
\vspace{-1em}
\end{table}

%% file: nature_sections/methodology.tex
\section{\name{} Framework}

\subsection{System Design}

The development of \name{}, an AI-driven system for generating situation reports, followed an iterative human-centered design approach. The initial phase focused on designing and developing the backend workflow, along with a preliminary frontend interface, and included multiple evaluation stages. Throughout this process, numerous presentations were made to stakeholders in both government and the private sector, which provided valuable feedback. The second phase involved engaging intelligence analysts and decision-makers through formative and collaborative design studies to better understand their needs and expectations.
The formative study phase (\S{\ref{sec:formative_study}}) involves semi-structured interviews to gather insights from users, involving detailed analysis of expectations from AI systems. Subsequently, the collaborative design phase (\S{\ref{sec:collaborative_design}}) brought users and developers together to refine and validate the initial design concepts. Through interactive sessions, participants provided real-time feedback on prototype functionalities to obtain precise requirements for AI assistance in report generation. The culmination of these efforts is a set of design strategies that ensure \name{} incorporates a user-centric design to serve the practical needs of intelligence analysts.

\subsubsection{Formative Study}
\label{sec:formative_study}

The formative study aims to gather information on the general needs and expectations from intelligence analysts for AI-driven systems. The study was conducted on ten intelligence analysts with experience in government and military roles (details on recruitment in Supplementary \S{\ref{sec:analyst_recruitment}}). Over a two-week period, semi-structured interviews were conducted with these analysts to examine their understanding, perspectives, and recommendations regarding AI use in professional settings.
The study highlighted emerging themes related to analysts’ perceptions and expectations
of AI-assisted authoring tools, which we briefly describe below:
\begin{itemize}
    \item \textit{Viewing technology as a means to enhance human capability}: An overwhelming majority (9 out of 10) emphasized the crucial role of AI in enhancing their capabilities, with these tools regarded not as mere process accelerators, but as essential elements that enrich their work by improving research efficiency, idea generation, and clarity of information. This perspective contrasts with the simplistic media depiction of these technologies as mere replacements for human effort.  
    
    \item \textit{Trusting and relying on machines, as with humans}: The majority of participants (8 out of 10) exhibited a tendency to attribute human-like qualities of trust and reliability to AI systems. The criteria for trusting AI closely resembled those for human interactions: the ability to provide dependable information, transparency in reasoning, and a foundation in verifiable facts. Interestingly, analysts did not set higher standards for AI than for human colleague.  This parity in trust and reliability criteria suggests that participants viewed AI as an equal collaborative partner, assessing its competence and trustworthiness on the same grounds as a human team member.
    
    \item \textit{Training and guiding AI}: Our study identified a split in intelligence analysts’ perspectives on their role in training and guiding AI systems. Four out of ten participants advocated for substantial control over AI, emphasizing the need for an interactive system that allows them to influence everything from information source selection to narrative shaping in reports. In contrast, the majority (six out of ten) favored a more hands-off approach, highlighting that situation report creation follows well-established, standardized procedures suitable for AI implementation. They perceived AI involvement as an extension of routine oversight, akin to reviewing a junior colleague's work.
\end{itemize}

\subsubsection{Collaborative Design}
\label{sec:collaborative_design}

To gain an operational understanding of intelligence analysis process and generating situation reports, we expanded the design opportunities identified in the formative study (\S\ref{sec:formative_study}) with subsequent collaborative design sessions with the ten analysts. The goal was to capture tangible design strategies and recommendations from users about how they, as intelligence analysts, navigate, research, and author their situation reports. 

\begin{figure}[t]
    \centering
    \includegraphics[width=1.0\textwidth]{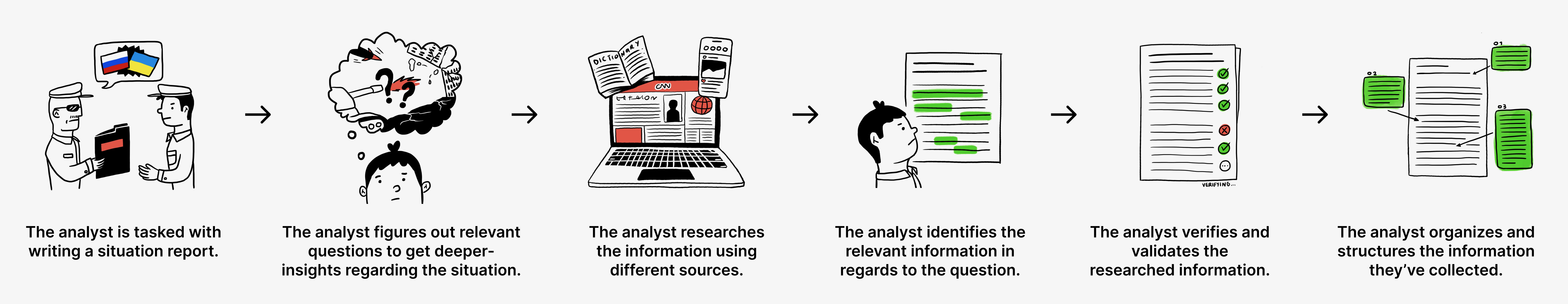}
    \caption{Storyboard used in the collaborative design sessions with intelligence analysts.}
    \label{fig:storyboard}
\end{figure}

We conducted study sessions with the analysts, with each session consisting a workflow review using storyboards and hands-on training with a simulation report exercise. In our study, participants engaged with a low-fidelity storyboard (shown in
Figure \ref{fig:storyboard}), where each panel depicted a distinct phase in situation report creation. Participants were tasked with providing detailed descriptions of each storyboard panel to ensure comprehension of the depicted scenario and workflow. Participants simulated each storyboard step using sample situations to gain practical workflow experience, and were advised to utilize diverse resources, including web search engines like Google and Bing, and Large Language Models (LLMs) such as ChatGPT~\cite{achiam2023gpt}, for task completion.

Data analysis from the collaborative design sessions showed three main themes: enhancing analytical efficiency, transparency in AI systems, and customization flexibility. Participants highlighted the need for interfaces that reflect their mental models, reducing cognitive load and allowing them to focus on strategic aspects. There was a significant emphasis on understanding AI systems' underlying logic for trust, with a preference for transparent methods and traceable data sources to verify the credibility of automated outputs. Additionally, users expressed a desire for tools that support varying analytical styles and complexities, and that can integrate information from diverse sources to provide a comprehensive analysis.
From the findings of the formative study and the collaborative design above, we identified the following design strategies:

\begin{itemize}
    \item \textbf{DS1:} Given the emphasis on reducing cognitive load and enhancing analytical efficiency, the system will be designed with an interface, that mirrors intelligence analysts' natural processes of data analysis and report generation.
    \item \textbf{DS2:} To increase efficiency, the system will integrate features to automate time-intensive tasks such as question curation and preliminary research, thereby reducing analysts' manual workload and enabling greater focus on strategic analysis and decision-making.
    \item \textbf{DS3:} The design, addressing the need for trust and reliability, will convey clear explanations of the system’s data processing algorithms and criteria. This includes transparent data sourcing, providing references within reports, and tools for users to easily understand and verify the system's conclusions. The design will also facilitate incremental trust-building through consistent and validated performance over time.
    \item \textbf{DS4:} Addressing the themes of customization and flexibility, the system will offer a high degree of adaptability to accommodate various analytical styles and levels of detail in reporting. It will include features for adjusting the depth of analysis, focusing on specific data sets, and seamlessly integrating various data sources.
\end{itemize}

\subsection{System Architecture}
\label{sec:smartbook_system}

\begin{figure}[t]
\centering
\includegraphics[width=1\linewidth]{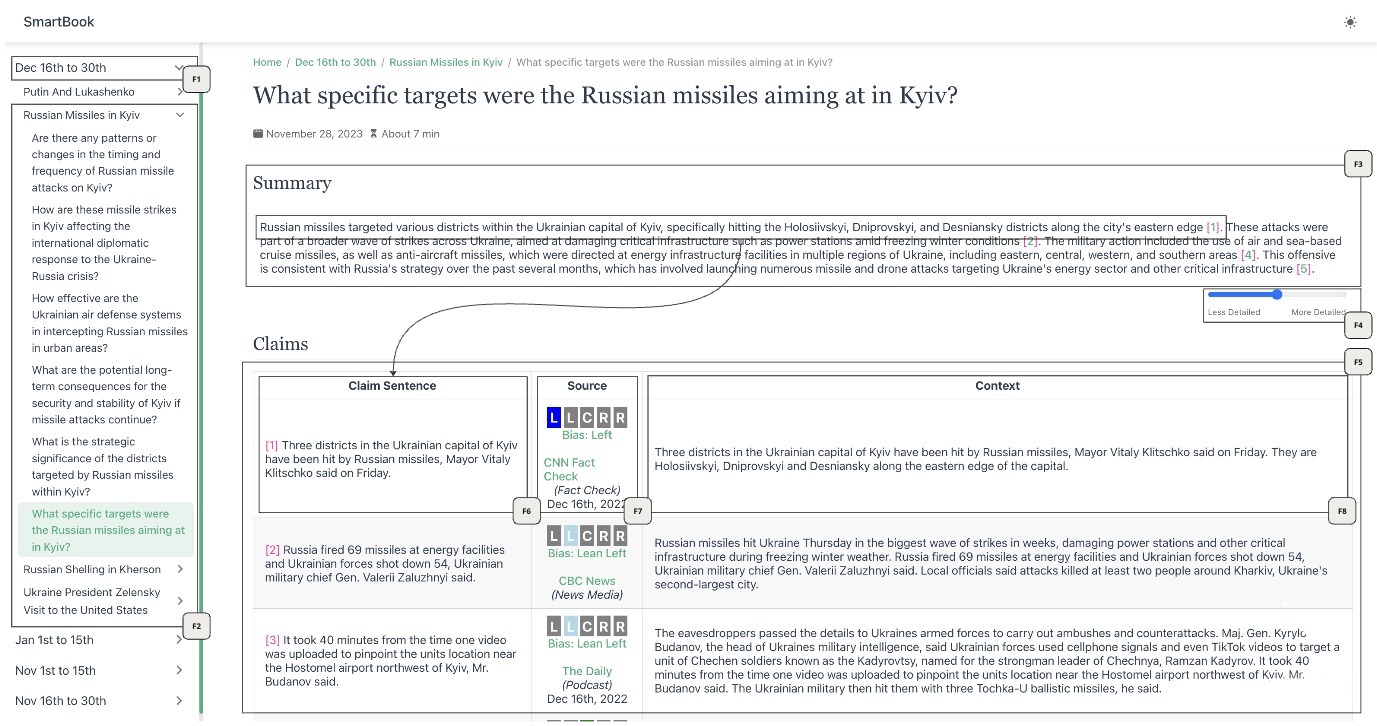}
\caption{A screenshot of \name{}'s front-end interface. Within the given situation, the user can navigate timelines (F1), explore strategic questions related to an event (F2), read the overarching summary on a given strategic question (F3), control the depth and length of information (F4), investigate all the claims in the summary (F5), trace each claim to corresponding summary fragment (F6), investigate the source metadata (F7) and read the context from which the claims were extracted (F8).}
\label{fig:system1}
\end{figure}

\begin{figure}
    \centering
    \includegraphics[width=0.95\linewidth]{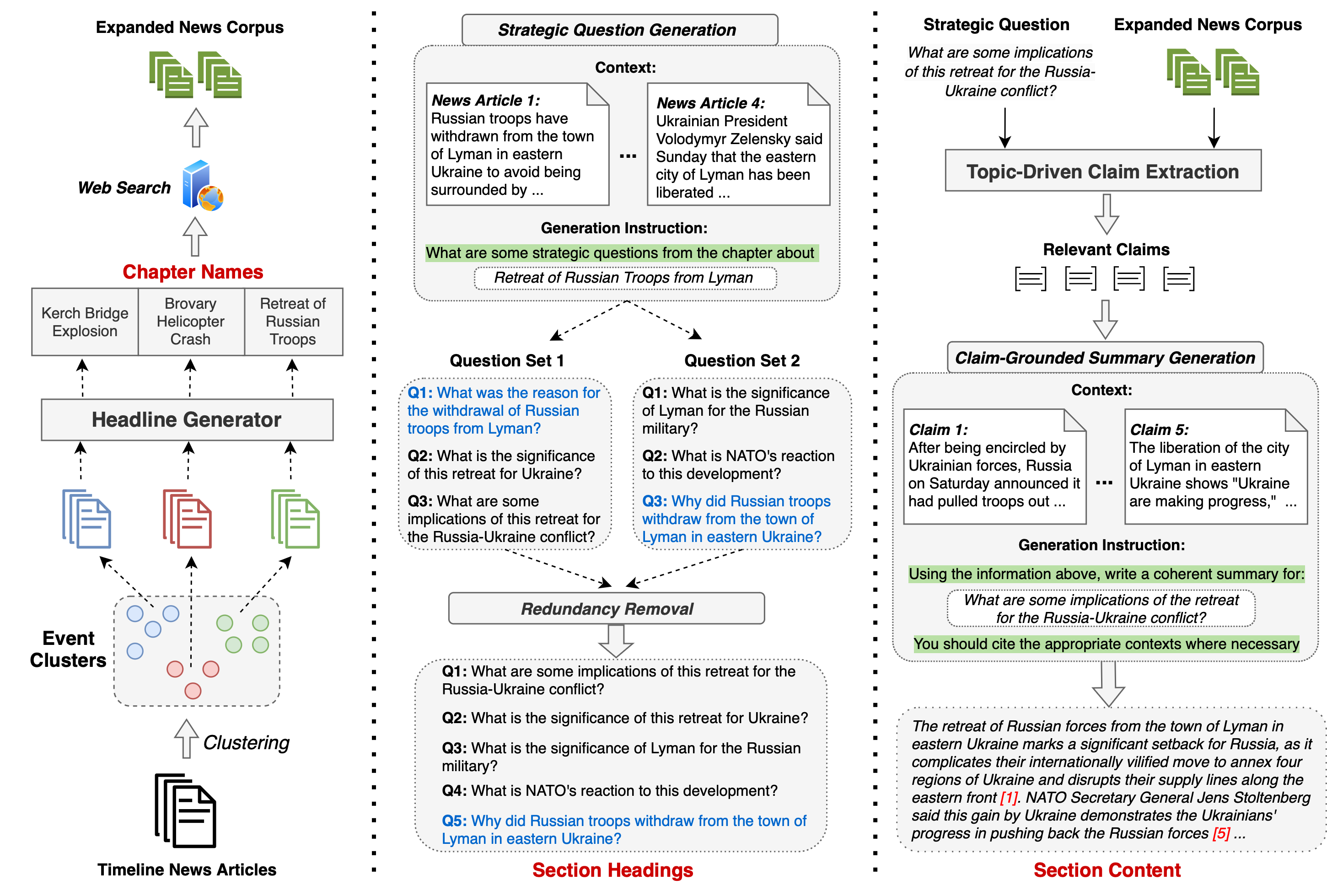}
    \vspace{-1em}
    \caption{\small Backend workflow for constructing \textit{\name{}}. Given the articles corresponding to a specific timeline, the figure shows the process for obtaining the chapters, their section headings, and the corresponding section content.}
    \label{fig:overall_workflow}
\end{figure}
The above four design strategies helped shape \name{}, an AI-assisted system for situation report generation that provides analysts with a first-draft report to work from as they respond to time-critical information requirements on emerging events.
\name{} consists of: 1) An intuitive user interface (shown in Figure \ref{fig:system1}) with design strategies from \S{\ref{sec:collaborative_design}}, and 2) a back-end framework (shown in Figure \ref{fig:overall_workflow}) that, when given a collection of documents from a variety of news sources, automatically generates a situation report.

Our automatic situation report is organized into coherent, chronological timelines spanning two weeks each, enhancing the tracking and comprehension of events' developments. Within these timelines, major events are identified by clustering news articles, forming the basis for subsequent chapters (see \S{\ref{subsec:major_event}}). To guide detailed chapter analysis, we incorporate a logical structure with automatically generated section headings in the form of strategic questions covering various aspects of each major event (details in \S{\ref{subsec:section_headings}}). \name{} generates content addressing these questions from a strategic perspective, by pinpointing relevant claims in news articles (see \S{\ref{subsec:claim_extraction}}). Each section contains query-focused summaries that answer the strategic questions, providing a comprehensive view of the event's contexts and implications (see \S{\ref{subsec:content_gen}}). These summaries include citation links that allow for factual verification and cross-checking by experts.


\name{} is designed for efficiency, utilizing smaller models~\cite{liu2019roberta, lewis2020bart} for tasks that have training data available, such as event headline generation, duplicate question detection, and claim extraction. Conversely, for more complex tasks—such as generating long-form summaries with citations or identifying strategic questions, we employ large language models~\cite{openai2021gpt35,achiam2023gpt}. 
In this section, we describe the various components within \name{}, along with emphasizing the advantages of each aspect of \name{}'s design for users (i.e., intelligence analysts) and for recipients of the final \name{} report (i.e., decision-makers), who both initiate information requirements and are downstream readers.

\subsubsection{Major Events within Timespans as Chapters}\label{subsec:major_event}

Situation reports cover event progressions over considerably long periods. Hence, it is beneficial to organize such reports in the form of timelines (F1 in Fig. \ref{fig:system1}), which enables seamless report updates~\cite{ma2023structured} with new events and helps facilitate~\cite{singh2016expedition} users tracking and understanding of situation context (informed by DS1). Timelines aid intelligence analysts in understanding event progression and predicting future trends by organizing events chronologically and highlighting cause-and-effect relationships. For readers, especially those less familiar with the subject, timelines provide a visual guide to easily grasp the sequence and significance of events in a scenario. Our automatic situation report has timelines to provide a coherent, chronological representation of event developments (DS1, DS2).

In each timespan, we first identify major events by clustering daily news summaries from the period into major event groups using an agglomerative hierarchical clustering algorithm~\cite{jain1988algorithms} based on term frequency-inverse document frequency (TF-IDF) scores~\cite{sparck1972statistical}. 
Finally, we are left with clusters of news snippets, each providing a focused view of a major event. However, since news summary snippets are condensed in detail, we improve the comprehensiveness of each event cluster by expanding the news corpus, as described next. First, to create a chapter name for \name{} and also use it for additional news article retrieval, we generate a concise headline for each event cluster. To achieve this, we utilize a sequence-to-sequence transformer-based \cite{vaswani2017attention} language model, BART \cite{lewis2020bart}, 
that takes the concatenated title and text from all the news snippets within the event cluster as input and generates a short event heading. We made use of the model trained on the NewsHead dataset \cite{headline2020} for generating multi-document headlines. In that dataset, each cluster contained up to five news articles, and a crowd-sourced headline of up to 35 characters described the major information in that cluster. We then use these chapter names as the query to retrieve additional relevant news articles via Google News.

\subsubsection{Strategic Questions as Section Headings}\label{subsec:section_headings}

A situation report should have a logical structure and descriptive section titles (F2 in Fig. \ref{fig:system1}) for clarity and easy access to information for intelligence analysts (DS2). \name{} not only describes event details in each chapter but also aims to present strategic insights that support decision-making and policy planning. To guide such detailed analysis, we incorporate a logical structure by automatically generating section headings in the form of strategic questions relating to each major event. These questions cover insightful details such as the motivations behind actions in an event and its potential future consequences.


Recent work \cite{sharma2021generative, wang2022towards} has shown that LLMs are capable of generating comprehensive, natural questions that require long-form and informative answers, in comparison to existing approaches \cite{murakhovska-etal-2022-mixqg, du-etal-2017-learning} that mainly generate questions designed for short and specific answers. In this work, we direct the LLM (GPT-4) to generate strategic questions about specific events, using news articles from the event cluster to anchor the context and reduce model-generated hallucinations~\cite{ji2022survey, maynez2020faithfulness}. 
To ensure diversity in the generated questions, we sample multiple question sets using nucleus sampling~\cite{holtzman2019curious}. 
Our analysis reveals that questions may occasionally be repeated across different sets, as depicted in Figure \ref{fig:overall_workflow} where duplicates are highlighted in \textcolor{blue}{blue}. To address this, we perform question de-duplication using a RoBERTa-large \cite{liu2019roberta} model trained on the Quora Duplicate Question Pairs dataset, thereby consolidating the sets into a singular, diverse collection of strategic questions relevant to the event. 

\subsubsection{Extraction of Claims and Hypotheses}\label{subsec:claim_extraction}
 Intelligence analysts, given the high stakes nature of their work but limited time, need systems that quickly identify key information in documents (DS2). This enables them to focus on urgent matters without sorting through irrelevant data. Hence, automated situation report generation should be able to identify and extract the most scenario-relevant and crucial information across multiple documents (F5 in Fig. \ref{fig:system1}). Readers of the situation report benefit from information salience because they are presented with a concise, relevant overview of a situation. Essential points need to be highlighted, to enhance readability and clarity. Moreover, we also present the bias of each news source (taken from \href{https://www.allsides.com}{AllSides}) to help analysts consider information presented from different perspectives.

Providing readers with a comprehensive understanding of event context requires foraging for different claims and hypotheses from the source documents (i.e., news articles) that help explain a situation \cite{toniolo2023human}. We adopt a Question Answering (QA) formulation to identify claims relevant to a given strategic question, driven by the ability of directed queries to systematically extract relevant claims from news articles, as demonstrated in recent research~\cite{reddy2022newsclaims, reddy2022zero}. 
Our QA pipeline utilizes a transformer-based RoBERTa-large encoder model \cite{liu2019roberta} that has been trained on SQuAD 2.0 ~\cite{rajpurkar2018know} and Natural Questions ~\cite{kwiatkowski2019natural}. The model takes as input the news corpus split into snippets along with the strategic question, and outputs answer extractions to these questions. The corresponding sentences that contain these answers are considered as the extracted claims. However, the risk of high-confidence false positives~\cite{chakravarti2021towards} necessitates validation~\cite{reddy2020answer, zhang2021joint} of these answers.
To address this, we utilize an answer sentence selection model~\cite{Garg_2020} that verifies each context against the strategic question. The model is a binary classifier with a RoBERTa-large backbone trained on datasets such as Natural Questions~\cite{kwiatkowski2019natural} and WikiQA~\cite{yang-etal-2015-wikiqa}, and outputs a validation score ranging from 0 (incorrect) to 1 (correct), which is used to select the top-5 relevant contexts for summarization.

\subsubsection{Grounded Summaries as Section Content}\label{subsec:content_gen}
Considering the issue of hallucination in LLM-based summarization~\cite{ji2023survey, li2023halueval, bang2023multitask}, factuality is far more important than creativity for situation report generation. A reliable situation report must be anchored in verifiable sources to ensure credibility (DS3). This supports analysts in drawing robust, evidence-based conclusions, and the embedded links act as a springboard to more extensive research for readers wanting to dive deeper (F6, F7, F8 in Fig. \ref{fig:system1}). Additionally, we offer summaries with varied detail levels—brief (2-3 sentences), standard (4-6 sentences), and extended (2 paragraphs)—to cater to different reader preferences (F4 in Fig. \ref{fig:system1}) (DS4).


Using strategic questions obtained for each chapter as section {\textit{headings}}, we incorporate query-focused summarization to generate each section's {\textit{content}}. A concise summary is generated for each section in \name{} based on the relevant claim contexts (F3 in Fig \ref{fig:system1}).
Recent work~\cite{goyal2022news, bhaskar2022zero, reddy2022sumren} has shown that humans overwhelmingly prefer summaries from prompt-based LLMs \cite{brown2020language, chowdhery2022palm} over models fine-tuned on article-summary pairs \cite{lewis2020bart, zhang2020pegasus, liu2022brio}, due to better controllability and easier extension to novel scenarios. For summary generation, we feed the LLM (GPT-4) with the top-5 most relevant contexts (from \S{\ref{subsec:claim_extraction}}) and instructions to summarize with respect to the given strategic question and include citations. This approach not only aids in maintaining accuracy by grounding on claim contexts but also enhances the trustworthiness of the summaries by allowing verification against the cited sources.

%% file: nature_sections/results.tex
\section{Results}

To comprehensively evaluate \name{} and its efficacy in aiding intelligence analysis, we performed three distinct but interrelated studies: a utility evaluation, a content review and an editing study. The utility evaluation~(in (\S{\ref{sec:utility_eval}}) aimed to answer three research questions concerning \name{}'s usability, focusing on its effectiveness for intelligence analysts and decision-makers in generating and interacting with situation reports. Complementing this, the content review~(in \S{\ref{sec:content_review_study}}) assessed the quality of \name{}'s automatically generated questions and summaries, focusing on readability, coherence, and relevance. Finally, the editing study~(in \S{\ref{sec:editing_study}}) involved an expert analyst editing the generated reports to assess their value as a preliminary first draft.

\subsection{Utility Study: Evaluating the System}
\label{sec:utility_eval}

During this user study, we conducted an interview with semi-structured questions and a post-study questionnaire on the usability of the \name{}. Each user study session was structured in four segments: (i)~an introductory overview, (ii)~a free-form investigation, (iii)~a guided exploration, and (iv)~a concluding reflective discussion. The participants were ten intelligence analysts (from \S{\ref{sec:formative_study}}) and two decision-makers from Canadian government boards. These decision-makers engaged in the initial qualitative study but did not answer subsequent post-study questionnaires, due to time constraints. The study investigated the following research questions:
\begin{itemize}
    \item \textbf{RQ1:} How do intelligence analysts interact and leverage the features within \name{}?
    \item \textbf{RQ2:} Do intelligence analysts find \name{} intuitive, usable, trustable and useful?
    \item \textbf{RQ3:} How do decision-makers interact, perceive and use \name{}?
\end{itemize}

 Upon commencement of the study,~(i) participants received a concise introduction to \name{}, outlining its core premise. This orientation was designed to acquaint them with the system without biasing their exploration. Subsequently,~(ii) participants were invited to freely investigate \name{}, with the specific task of exploring a minimum of three questions across five chapters of their choosing. This approach afforded substantial freedom, enabling interactions with the user interface that reflected their natural inclinations and interests. Following the free-form investigation, participants were systematically introduced~(iii) to the \name{} framework. This process involved tracing the journey from a chosen question related to a specific event within a designated timespan, to the corresponding claims, contexts, and sources, resulting in a summarized answer. Subsequently, a semi-structured interview~(iv) was conducted to gather reflective feedback on the participants' experience with \name{}, focusing on its efficacy and potential areas of enhancement. Finally, to conclude the session, participants were asked to complete a post-study questionnaire. This questionnaire was focused on assessing the usability of \name{}, gathering quantitative data to complement the qualitative insights gained from the semi-structured interviews. Upon collecting, discussing and iterating on the data, behaviors and insights were merged into the following themes:

 \begin{figure}[t]
    \centering
    \includegraphics[width=1.0\textwidth]{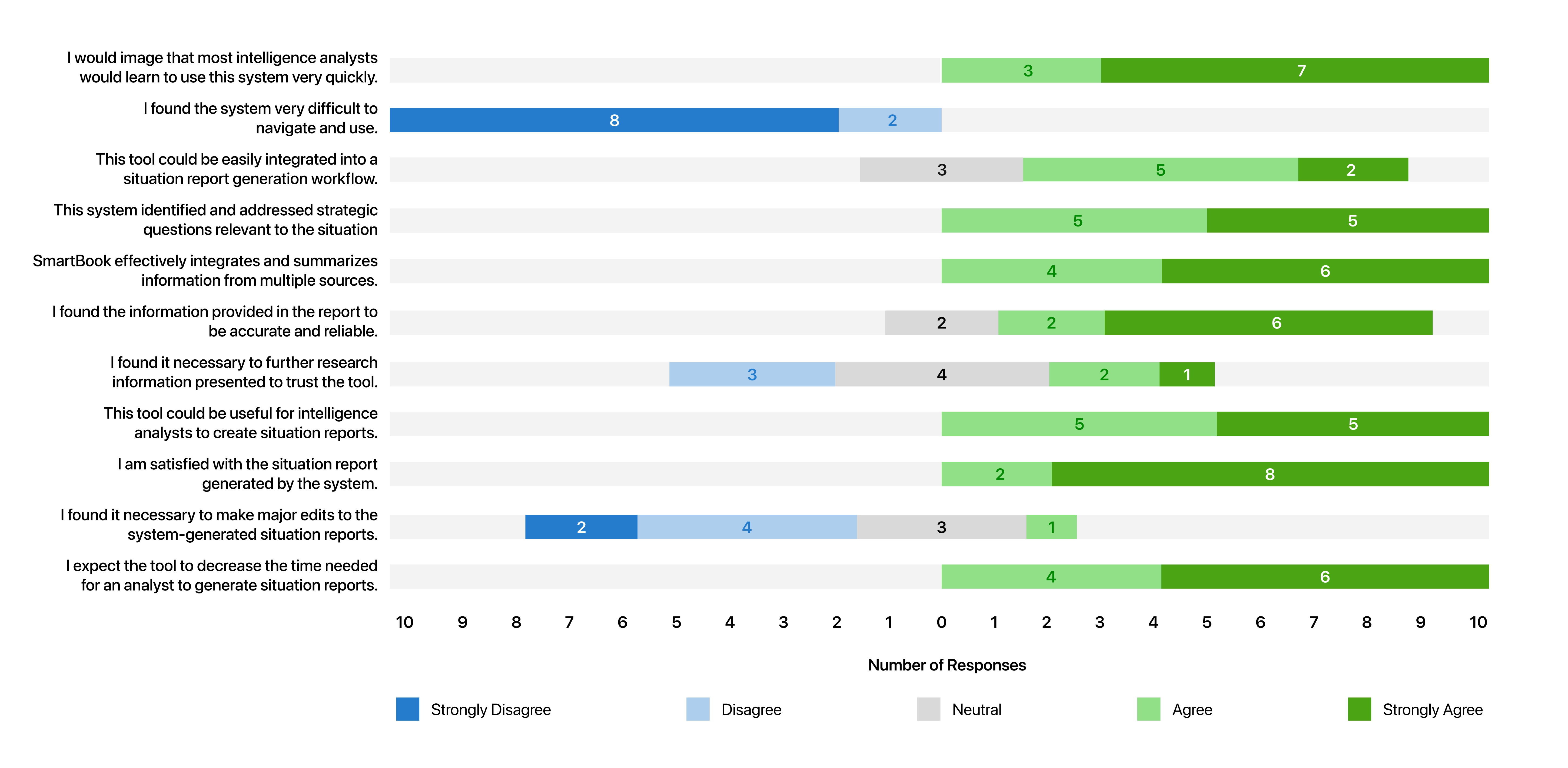}
    \vspace{-3em}
    \caption{Quantitative results from the post-study questionnaire in the utility evaluation.}
    \label{fig:userstudy}
\end{figure}

\subsubsection{UI Understandability and Interaction}
The structured layout of the user interface (UI) was notably effective, as seen in the free-form investigation~(ii) with a 100\% feature discovery rate (10 out of 10 participants). All participants
successfully engaged with key features, underscoring the UI's design intuitiveness. This high feature discovery rate indicates that the UI aligns with users' cognitive patterns, facilitating
intuitive interaction without extensive training, providing us with a positive response to \textbf{RQ1}. This sentiment was further emphasized in the post-study questionnaire. As seen in the results in Figure~\ref{fig:userstudy}, 70\% of participants strongly agreed that most intelligence analysts would learn to use the system very quickly, while 80\% strongly disagreed that the system was difficult to navigate and use. 
    However, we observed mixed views on UI flexibility, specifically in personalization and customization. While most found the tool easily integrateable into their workflow, a minority (30\%) remained neutral. When further prompted, two participants described the UI as a ``one size fits all'' solution, lacking personalization.

\subsubsection{Building Trust on \name{}} 
In the study, trust in the \name{} system developed progressively, resembling the formation of trust in a human analyst. Initially, users exhibited skepticism, thoroughly scrutinizing the system's sources and assertions. This included validating the authenticity of sources, the accuracy and pertinence of ratings, and the correct representation of context. As users became more acquainted with \name{} and evaluated its reliability, their dependence on extensive source verification lessened. As can be seen in Figure \ref{fig:userstudy}, three out of ten participants did not feel the need to conduct additional research beyond the presented information to trust the tool, while another three felt additional research was necessary. This trust was found to be context-dependent, with two participants noting that their trust varied based on the ``impact severity'' or the potential negative consequences of disseminating incorrect information. Nonetheless, a majority (eight out of ten) concurred that the information provided was accurate and reliable, a positive answer to \textbf{RQ2}. 

\subsubsection{Perceived Benefits for Intelligence Analysts}
The primary benefit identified was the substantial reduction in time and effort required for compiling and analyzing complex data. \name{}'s automation of initial report generation processes, such as data collection and summarization, was highly valued by analysts as it allowed more focus on thorough analysis and strategic planning. As seen in Figure \ref{fig:userstudy}, all participants agreed on the tool's utility in assisting intelligence analysts in creating situation reports, expressing satisfaction with the system-generated reports, thereby providing a positive response to \textbf{RQ2}. Regarding the necessity of significant edits to the system-generated reports, half participants suggested that this need was not inherently due to report deficiencies but varied based on the report's intended purpose or audience. We later conducted a study (described in \S{\ref{sec:editing_study}}) to understand the extent of edits needed.

\subsubsection{\name{} as a Learning Tool for Decision-Makers}
Decision-makers highly valued \name{}'s ability to rapidly deliver accurate and easily digestible information. Their primary engagement with the interface centered around utilizing the summaries, to learn about different topics in various degrees. The system's effectiveness in simplifying complex data into structured, clear formats was particularly appreciated, to help aid in swift understanding and decision-making processes. Although data lineage and source transparency were recognized as important, these were considered secondary to the primary need for timely and format-specific information delivery. While \name{}'s target users are intelligence analysts, the decision-makers' view highlights \name{}'s dual functionality as both an analytical tool and a decision-support system, providing a key capability for the high-paced and information-intensive needs of decision-makers, addressing \textbf{RQ3}.

\subsection{Content Review: Evaluating the Generated Content}
\label{sec:content_review_study}

Our content review study primarily involves evaluation of the quality of automatically generated strategic questions and summaries within \name{}. The studies involved two participant groups: one senior expert intelligence analyst and six text evaluators. The expert analyst, affiliated with a US national defense and security firm, possessed a decade of experience in intelligence analysis, training, and AI-based tool assessment. The text evaluators, all fluent in English and US-based, involved undergraduate and graduate students with experience in Natural Language Processing.

\subsubsection{Evaluation of Strategic Questions}

Evaluation of the quality of strategic questions, which are used as section headings in \name{}, is done by drawing parallels to the traditional construction process of a situation report, which involves the participation of
both senior and junior analysts. A senior analyst’s role is usually to come up with strategic questions, which are then passed onto a junior analyst. The junior analyst then gathers tactical information that can help answer or provide more background to the strategic questions. Our human evaluation of the questions within \name{} measures the following aspects, based on guidelines defined by the expert intelligence analyst:
\begin{itemize}
    \item \textbf{Strategic Importance}: Evaluating the strategic importance of a question requires introspection on whether the
question provided within \name{} probes for an insightful aspect of the event. The scoring rubric involves marking the question as one of three categories, namely, `Not Strategic', `Some Strategic Value', and `Definitely Strategic'. 
This metric is based on the level of speculative nature, such as questioning the background/motives/reasons behind actions taken by actors (governments/military, etc) involved in the event.
    \item \textbf{Tactical Information}: Another dimension of evaluating the quality of a question is how much relevant tactical information can be gathered using that question. In this case, the question is evaluated based on its corresponding \name{} summary, in terms of whether the question-relevant tactical information in the summary is helpful for an analyst to gain deeper insights into the situation.  Tactical information is defined as content that is neither obvious nor trivially obtained (e.g. cannot be obtained from a news header).  The scoring rubric is categorized as `No information is tactical', `Some information is tactical', and `Most information is tactical'. Table \ref{tab:tactical_info_examples} in supplementary shows examples for summaries with corresponding tactical information.
\end{itemize}

We randomly selected 25 chapters from \name{}, comprising 125 strategic questions, with each question evaluated by three annotators. Results show that most questions are strategic (detailed split in Figure \ref{fig:question_eval} in supplementary), with at least 82\% having some
strategic value. Further, we see that these questions can help gather relevant tactical information in roughly 93\% of the cases, as judged by the annotators that most summaries to these questions have tactically relevant information. When assessing the diversity of strategic questions within each chapter, evaluators noted that 64\% of chapters contained questions addressing different aspects, 28\% had up to two questions on similar aspects, and 8\% included more than two questions on similar aspects.

\subsubsection{Evaluation of \name{} Summaries}

Next, we assess the quality of text summaries generated by \name{}, which uses question-driven claim extraction to identify and summarize relevant information. For baseline comparisons, we utilize two methods: firstly, a \textit{query-focused summarization} baseline that processes entire news articles directly with an LLM to generate summaries for the strategic question as the query. This approach uses the same LLM and prompt as in \name{}, except the entire news article texts are passed as input context without an explicit claim extraction step. Secondly, we compare against a web search + LLM baseline, where relevant web pages sourced from the internet serve as the input context, which are then summarized by an LLM. This method simulates using an LLM-enabled web search engine (like \href{https://www.perplexity.ai}{perplexity.ai}), incorporating the strategic question as the query along with a phrase ``\textit{concerning the Ukraine-Russia crisis between $<$timeline$>$}''. Since the LLM-enabled web search was using GPT-3.5~\cite{openai2021gpt35} at the time of experiments, we used GPT-3.5 as the LLM in SmartBook and other baselines for a fair comparison.
We randomly select summaries corresponding to 50 strategic questions from \name{} to perform the evaluation.
\input{tables/sum_human}

\input{tables/summaries_comparison}
For each generated summary, three evaluators are asked to assess the coherence, relevance, and usefulness with scores ranging from 1 (worst) to 5 (best). Simply, \textit{coherence} measures the quality of all summary sentences collectively regardless of the question. \textit{Relevance} quantifies whether the summary contains key points that answer the given question. We define \textit{usefulness} as an indication of whether the summary provides non-trivial and insightful information for analysts, and suggests the breadth and depth of the provided key points. 
Table~\ref{tab:sum_human} shows results from the evaluation study. We observe that \name{} outperforms alternative competitive strategies in coherence, relevance, and usefulness. The advantage of the question-driven claim extraction in \name{} is evident, yielding significantly more relevant summaries compared to the direct query-focused summarization without such a step (row (3) vs (1)). Additionally, summaries generated using information sourced from the Web are less useful than those generated through \name{}'s news-driven approach (row (2) vs (3)). Table \ref{tab:summaries_comparison} displays an example showing the difference in outputs. Overall, the evaluation demonstrates that across the metrics, \name{} excels over the baselines in providing high-quality summaries.


We also evaluated the citation quality in \name{}'s summaries using metrics for citation precision and recall as defined by \citet{gao2023enabling}. Citation recall assesses whether the output sentence is fully supported by the cited context, whereas citation precision identifies any irrelevant citations. These metrics were calculated using the 11B-parameter TRUE model \cite{honovich-etal-2022-true-evaluating}, which is trained on a collection of natural language inference datasets, and is frequently employed to evaluate attribution accuracy~\cite{bohnet2022attributed, gao2023rarr}.  Overall, 97\% of the sentences in \name{}'s summaries included citations, while 29.5\% had multiple citations. We observed a citation precision and recall of 64.7\% and 69.2\%, respectively.


\subsection{Editing Study: Evaluation as a Preliminary Draft}
\label{sec:editing_study}
\name{} serves as an initial draft for intelligence analysts to refine or adapt to their specific needs.  We evaluate \name{}'s effectiveness in producing situation reports through an editing study with an expert analyst, by measuring how much of the content the analyst directly accepts or further edits. The analyst actively explored \textsc{SmartBook}'s features and subsequently edited 94 randomly selected summaries from \name{}-generated situation reports, until they met professional intelligence reporting standards. 

  Based on the revised summaries by the expert analyst, we assessed the alterations made to the original content. We quantified these changes using token-overlap metrics like BLEU~\cite{papineni2002bleu} and ROUGE~\cite{lin2004rouge} scores, and Levenshtein edit distance, which calculates character-level modifications needed to transform the original into the revised summary.
Empirical results show high token overlap between the generated and post-edited texts, with BLEU and ROUGE-L scores respectively at 59.0\% and 74.1\%, indicating that the \name{}-generated reports are of sufficiently good quality and extensive human expert revision may not be necessary. However, we acknowledge that a gap still exists between the automatically generated summaries and human expert curation, as the Levenshtein edit distance computed at the character level is 34.4\%. Notably, 15\% of the generated summaries had no edits made by the expert analyst.

Further analysis revealed that the analyst predominantly added rather than removed content, with insertions at 49.6\% and deletions at 2.3\%. This suggests that automated summary generation may generally need to be more detailed.  
\input{tables/table2}
Table \ref{tab:human_edits} shows an example of edits (in color) made by an expert analyst for a machine-generated summary in \name{}. We can see here that the human analyst added additional tactical information (in blue) to elaborate on certain aspects (e.g. what is special about the ``kamikaze'' type of drone). Further, the analystalso added some interesting insights (in green) based on the information in the summary. Overall, this shows that \textsc{SmartBook} provides a good starting point for analysts to expand upon for the generation of situation reports. 

It is noteworthy that 15\% of summaries produced by \name{} needed no modifications, highlighting its proficiency in creating acceptable reports in some scenarios without human intervention. This suggests that as technology advances and iterative refinements are applied, this rate will likely improve, reducing the workload for analysts in the future. 
To gain a better understanding of the different types of errors in the remaining summaries, we asked the expert analyst to categorize the errors within them. The analyst was also shown the strategic question and the corresponding extracted contexts that were used to automatically generate the summary. The summary errors were categorized as follows:

\begin{itemize}
    \item \textit{No relevant contexts:} None of the extracted contexts are relevant to the question (and thereby the summary is expected to be irrelevant too).
    \item \textit{Inaccurate information in summary:} Summary has incorrect information, that is not reflective of the underlying input contexts.
    \item \textit{Incoherent summary:} Summary is incomprehensible and unclear.
    \item \textit{Incomplete summary:} Important information in the input contexts is missing in the summary.
    \item \textit{Irrelevant information in summary:} Summary has material that is not relevant to the question, despite some extracted contexts being relevant.
\end{itemize}

\begin{wrapfigure}{r}{0.3\linewidth}
\vspace{-1.65em} 
        \centering
        \includegraphics[width=0.9\linewidth]{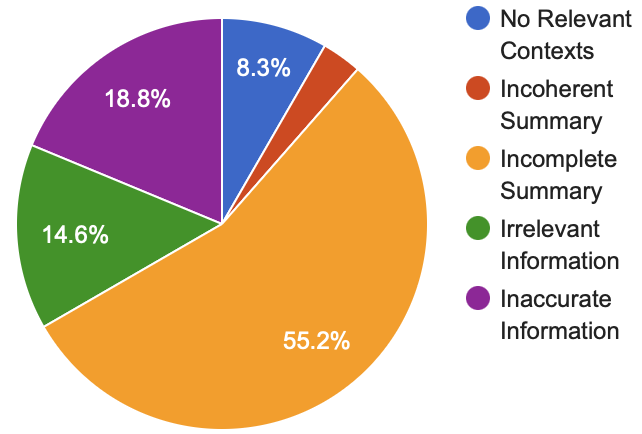}
        \vspace{-0.5em} 
        \caption{Distribution of different error types for summaries within \name{}.}
        \label{fig:error_cat} 
        \vspace{-1em} 
\end{wrapfigure}

Figure \ref{fig:error_cat} shows the distribution of error categories for the summaries. It can be seen that incompleteness of summaries is a predominant error, with more than 50\% of the summaries missing important information or not being sufficiently complete.  While the predominance of incomplete summaries could be a concern, this can also be framed positively: it represents a conservative approach ensuring that \textsc{SmartBook} does not over-extend based on limited data, and instead offers a foundational understanding.  Other errors corresponded to summaries with inaccurate (18.6\%) or irrelevant (14.6\%) information. Directing the LLMs to reference input documents while generating summaries does improve factuality, aligning with recent findings~\cite{gao-etal-2023-enabling}. Nevertheless, LLMs' tendency to hallucinate~\cite{ji2022survey, tam2022evaluating}, calls for validation and cross-referencing techniques (discussed in \S{\ref{sec:future_extensions}}) to address these issues. Finally, we see that very few summaries were judged incoherent, as expected given that large language models have been shown \cite{goyal2022news, zhang2023benchmarking} to generate fluent and easy-to-read output.

%% file: tables/sum_human.tex
\begin{table}
\centering
\begin{tabular}{lcccc}
\hline
\textit{Summary-Type} & \textit{Background} &  \textit{Coherence} & \textit{Relevance}  &\textit{Usefulness} \\
\hline

(1) Query-Focused & News & 4.39& 3.85& 3.77\\
(2) Web Search + LLM & Web &4.44 & 4.07& 3.83\\
(3) SmartBook & News & \textbf{4.56} & \textbf{4.23} & \textbf{4.17}\\

\hline
\end{tabular}
\caption{Results of human evaluation comparing the summaries within SmartBook (3) against those either generated directly from news articles via query-focused summarization (1) or obtained from the internet by web search combined with generation from a large language model (2). Results are on a 1-5 scale (worst to best). 50 summaries were evaluated with a text evaluator coverage of 3.}
\label{tab:sum_human}
\vspace{-2em}
\end{table}

%% file: tables/summaries_comparison.tex
\begin{table}[t]
    \centering
    \small
    \renewcommand{\arraystretch}{1.5}
    \begin{tabular}{p{17.9em}|p{14.3em}|p{17.4em}}
    \multicolumn{3}{c}{\textbf{Question:} How have kamikaze drones changed the nature of the Russia-Ukraine conflict?} \\
    \hline
     \multicolumn{1}{c|}{\centering \textbf{Query-Focused}}  & \multicolumn{1}{c|}{\centering \textbf{Web Search + LLM}}  &\multicolumn{1}{c}{\centering \textbf{SmartBook}}\\
     \hline
    Kamikaze drones, or suicide drones, have been used by Russia in its attacks on Ukrainian cities such as Kyiv, Vinnytsia and Odesa. They are known as a ``loitering munition'' because they are \textcolor{blue}{capable of waiting in an area identified as a potential target and only strike once an enemy asset is identified}. Russia is using Iranian-made Shahed-136 drones, which have a payload of approximately 50 kilograms and are \textcolor{blue}{hard to detect due to their small size}. Ukraine has responded by \textcolor{blue}{developing an app that civilians can use to report sightings of incoming drones} and missiles, and increasing the proportion shot down before they hit the ground. &  Russia \textcolor{red}{and Ukraine} have been using Iranian-made Shahed-136 drones in the conflict. These drones are known as `kamikaze' drones due to their \textcolor{blue}{ability to self-destruct on impact}. The use of these drones has been highlighted by \textcolor{blue}{Russia's growing reliance on them and their use as both military weapons and psychological weapons}. On October 13, 2022, kamikaze drones and missiles hit north and south Ukraine, though the Kremlin denied any involvement from President Putin in talks between Russia and the West. & The Russia-Ukraine conflict has seen a \textcolor{blue}{paradigm shift in the nature of Russian tactics, with the increased reliance of kamikaze drones over artillery fire}. These drones, which are \textcolor{blue}{designed to crash into a target and explode, can spray potentially deadly debris} and are \textcolor{blue}{being used to plug gaps in the deployment of Russian forces along the front line}. Russia is reportedly using Iranian-made Shahed series drones, which are showcased at Kashan Airfield south of Tehran. This shift in tactics has seen an \textcolor{blue}{escalation in the aerial bombardment of densely populated urban areas and an attack on critical infrastructure and power sources}.\\
       \hline
    \end{tabular}
    \caption{An example comparing the summaries produced by different approaches evaluated in Table \ref{tab:sum_human}. It can be seen that the SmartBook summary contains considerably more question-relevant information (highlighted in \textcolor{blue}{blue}). The \textit{Web Search + LLM} baseline output contains hallucinated content (highlighted in \textcolor{red}{red}) that suggests \textit{both} Russia and Ukraine have been using the drones.}
    \label{tab:summaries_comparison}
    \vspace{-2.5em}
\end{table}

%% file: tables/table2.tex
\begin{table}[t]
    \centering
    \small
    \renewcommand{\arraystretch}{1.5}
    \begin{tabular}{p{53em}}
      \hline
      \textbf{SmartBook Summary}: Russia has reportedly stepped up its use of kamikaze drones in its assault against Ukraine. The increased reliance on kamikaze drones over artillery fire likely signals a paradigm shift in Russian tactics a shift introduced to counter high mobility offensive probing by Ukrainian forces.  \\
      \hline
      \textbf{Analyst-edited Summary}: Russia has reportedly stepped up its use of kamikaze drones in its assault against Ukraine. \textcolor{blue}{The aircraft are called kamikaze drones because they attack once and don’t come back.} The increased reliance on kamikaze drones over artillery fire likely signals a paradigm shift in Russian tactics - a shift introduced to counter high mobility offensive probing by Ukrainian forces. \textcolor{blue}{Their low price means the drones can be deployed in large numbers and they hover before they strike, so they have a psychological effect on civilians as they watch and wait for them to strike. These drones allow Russia to target Ukrainians far away from the front line, away from the primary battle space.} \textcolor{green}{The emergence of swarms of drones in Ukraine is part of a shift in the nature of the Russian offensive, which some speculate indicates that Moscow may be running low on long-range missiles.} \\
      \hline
    \end{tabular}
    \caption{Example showing expert analyst edits for an automatic machine generated SmartBook summary on the use of `kamikaze' drones in the Ukraine-Russia crisis. Text that has been added by the expert is colored, with \textcolor{blue}{blue} corresponding to additional tactical information, whereas \textcolor{green}{green} corresponds to insights/conclusions added by the analyst.}
    \label{tab:human_edits}
    \vspace{-2em}
\end{table}

%% file: nature_sections/discussion.tex
\subsection{Discussion}


Amid  advancements in AI-assisted writing tools~\cite{WangACL2019,Wang2020,selfcollaboration2023,cardon2023challenges} tailored for diverse end-users, including academics~\cite{gero2022sparks}, screenwriters~\cite{mirowski2023co} and developers~\cite{chen2021evaluating}, \name{} introduces a specialized automated framework for generating situation reports for intelligence analysts. Unlike previous systems such as CoAuthor~\cite{lee2022coauthor}, Creative Help~\cite{roemmele2015creative}, Writing Buddy~\cite{samuel2016design} and WordCraft~\cite{yuan2022wordcraft}, which support collaborative and creative tasks, \name{} focuses on factual accuracy, analytical depth, and efficiency in structured, data-driven tasks. This distinct approach emphasizes less creative interaction and more rigorous information processing, aligning \name{} closely with the needs of intelligence analysis. However, the design also stands to benefit from iterative refinement (discussed in \S{\ref{sec:future_extensions}}), exploring deeper integration into analysts' workflows.


\name{}, with its modular design, automates the generation of situation reports efficiently across various contexts, including geopolitical, environmental, or humanitarian situations. 
Unlike traditional approaches that require substantial research and domain-specific expertise, \name{} can adapt to new domains with minimal configuration, thereby delivering accurate reports in diverse and rapidly changing global contexts. \name{}'s efficacy was primarily evaluated on news from the Ukraine-Russia military conflict. Here, we briefly explore \name{}'s applicability in a humanitarian scenario--the Turkey-Syria earthquake. 
By modifying the input data to identify key events in the specified time window (February 6-13, 2023), \name{} produced relevant reports detailing the earthquake’s aftermath (\href{https://www.cnn.com/middleeast/live-news/turkey-syria-earthquake-updates-2-7-23-intl/h_a9e1d082dd54c67187fd51322f95eaf2}{link}), international aid (\href{https://www.incirlik.af.mil/News/Article-Display/Article/3292036/urban-search-and-rescue-teams-arrive-at-incirlik-air-base/}{link}), rising casualties (\href{https://www.reuters.com/world/middle-east/death-toll-syria-turkey-quake-rises-more-than-8700-2023-02-08/}{link}), and notable incidents like the disappearance of a Ghanaian soccer star (\href{https://www.marketwatch.com/story/soccer-star-christian-atsu-still-missing-after-turkey-earthquakes-i-still-pray-and-believe-that-hes-alive-says-partner-cfd47272}{link}).
This adaptability from military to humanitarian crises highlights \name{}’s robustness and versatility. It also emphasizes the importance of the source articles and suggests that, with mission-relevant data input, \name{} has the potential to be an invaluable tool for analysts across diverse scenarios.

\subsubsection{Outlook}
\label{sec:outlook}

We have developed \textsc{SmartBook}, an innovative framework that generates situation reports with comprehensive, current, and factually-grounded information extracted from diverse sources. \name{} goes beyond mere information aggregation; it presents chronologically ordered sequences of topically summarized events extracted from news sources and placed into a UI layout structure that aligns with the workflow of intelligence analysis. Throughout the development of \textsc{SmartBook}, from its conceptual design to its evaluation, we engaged analysts to ensure the tool meets their practical needs and enhances their workflow. Our formative study and collaborative design efforts have been guided by the needs of intelligence analysts, particularly in addressing the challenge of data overload. The findings from these studies indicate that analysts are cautiously optimistic about integrating AI assistance into their processes, expressing a clear desire to shape a future in which AI tools can both adapt to the evolving data landscape and personalize to their individual analytical techniques.


\name{} establishes a foundation for situation understanding in a variety of scenarios. However, history repeats itself, sometimes in a bad way. Historical patterns, particularly those leading to natural or man-made disasters, can be a useful signal to trigger proactive measures for crisis mitigation.  
Building on prior research in event prediction~\cite{li2021future, wang2022schema, li2023open}, \name{} can guide the development of a news simulator that forecasts event outcomes in crisis situations.
This tool would be invaluable for humanitarian workers and policymakers to exercise reality checks, enhancing their capacity to prevent and respond to disasters effectively.

\subsubsection{Limitations and Future Extensions}
\label{sec:future_extensions}

While \name{} represents a significant advancement in the automated generation of situation reports, it is essential to acknowledge certain limitations that stem from both the technical aspects of the system and the scope of its application. These include (i) unverified news source credibility, (ii) potential inaccuracies in reflecting source material despite citations, and (iii) user studies focused mainly on military intelligence analysts, which may not represent the needs of a wider analytical audience. Recognizing these limitations is crucial for guiding future improvements and ensuring the framework's applicability across diverse intelligence analysis sectors. We elaborate on the limitations in more detail in the supplementary material. Here, we provide key future extensions designed to elevate \name{}'s utility as a comprehensive, unbiased, and reliable tool for situation report generation in intelligence analysis:

\begin{itemize}
    \item \textit{Incorporating Multimodal, Multilingual Information}: 
    Intelligence analysis increasingly relies on integrating diverse data types such as text, images, videos, and audio to understand global events comprehensively. We aim to enhance \name{}'s situation reports by correlating textual claims with corresponding multimedia, utilizing advanced systems like GAIA~\cite{li2020gaia} for multimedia knowledge extraction. 
    Furthermore, global events' reach necessitates multilingual intelligence analysis to capture local nuances missed by solely using English. Employing cross-lingual techniques~\cite{du2022resin} allows for the inclusion of multiple languages, thereby enhancing understanding of local dynamics and cultural subtleties that influence global situations. Thereby, integrating multilingual sources democratizes intelligence analysis, shifting away from a predominantly English-centric perspective. Ultimately, \name{} aims to set new standards in intelligence reporting, fostering a more inclusive and globally informed approach. \\
    \item \textit{Controlling the Bias of News Sources}: 
News sources carry inherent biases influenced by editorial policies, audience demographics, and geographical locations. Dominance of a single perspective may obscure crucial details or alternative viewpoints. In developing the next version of \name{}, we seek to mitigate these biases by incorporating a diverse spectrum of news outlets representing various political stances. This approach aims to diminish informational blind spots and expand the scope of scenarios available to intelligence analysts, thereby enhancing \name{}'s utility.\\
    \item \textit{Co-Authoring with Iterative Refinement}: 
    Authoring situation reports is an iterative process, where analysts continuously refine and update the reports. In the next iteration, we aim to provide a dynamic multi-turn editing process with \name{}, by leveraging analyst feedback for self-improvement. Techniques like Reinforcement Learning with Human Feedback~\cite{stiennon2020learning, bai2022training} and other personalization algorithms~\cite{wu2019npa, monzer2020user} will further enhance the system's capability to integrate human feedback, and progressively adapt to the preferences and decision-making styles of analysts. This strategy will align \name{} more closely with human analysts, leading to tailored situation reports that meet the specific requirements of intelligence analysis workflows. Ultimately, \name{} will evolve from a preliminary drafting tool to a comprehensive AI co-author for generating situation reports.\\
    \item \textit{Improving Reliability of the Generated Reports}: The integrity of data and claims in situation reports is crucial for strategic decision-making. A key challenge in automating these reports is guaranteeing the accuracy and reliability of data from diverse sources. To address this, we propose introducing a `verification score' for each claim, which evaluates the reliability of information based on source credibility, corroborative data, and historical accuracy of similar claims. 
    This mechanism provides intelligence analysts with a confidence metric to quickly assess data reliability. Higher scores indicate greater confidence and facilitate rapid integration into reports, whereas lower scores necessitate comprehensive review and possibly further verification.
\end{itemize}

%% file: nature_sections/supplementary.tex
\section{Supplementary Material}

\subsection{Recruitment Details}
\label{sec:analyst_recruitment}

For our studies, we targeted individuals with experience in government and military roles. We distributed a pre-screening survey to ascertain their background in creating situation reports. Participants qualifying for the study were either experienced intelligence analysts or had a minimum of one year of equivalent experience. The final group comprised 10 military personnel from different branches, as shown in Table \ref{tab:analyst_demographic}. Their experience in intelligence report writing varied between 1 and 10 years. Compensation ranged from \$25 to \$35 per hour,
reflecting participants’ levels of experience.

We implemented a targeted outreach strategy to recruit decision-makers for our study. These participants, identified through email addresses sourced from government boards and social media profiles (e.g., LinkedIn, Twitter), were self-identified professionals who make decisions based on prepared reports in their official capacity. Our group comprised of individuals with current or past experience on Canadian government boards. 
Unlike intelligence analysts, these participants did not receive compensation due to their government affiliation.

\input{tables/intelligence_analyst_demographic}

\begin{figure}[b]
    \centering
    \includegraphics[width=0.8\linewidth]{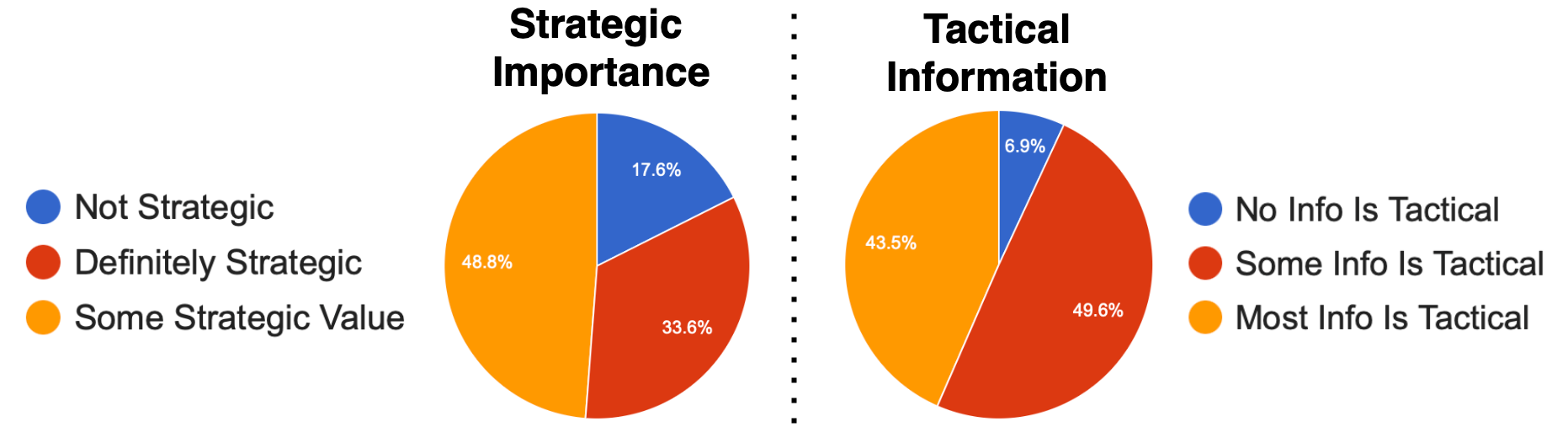}
    \caption{Results from the human evaluation of the strategic questions (section headings) in \name{}. Judgments are based on the strategic importance of the question and tactical information that can be gathered using the question.}
    \label{fig:question_eval}
\end{figure}

\begin{table}[!htb]
    \centering
    \renewcommand{\arraystretch}{1.5}
    \small
    \begin{tabular}{p{24.5em}|p{26.5em}}
      \textbf{Question:}  What were the motives behind the attack on Russian missile in Dnipro?
  &  \textbf{Question:} What is the capability of Ukraine's air defense systems in repelling the missile attacks from Russian missiles\\
  \hline
    \textbf{Summary:} The Russian missile attack on the city of Dnipro on Saturday was \textcolor{blue}{part of the Kremlin's efforts to limit Ukraine's ability to heat and power itself in the middle of winter}. The attack was reportedly \textcolor{red}{carried out with a Kh-22 missile}, an older and less accurate weapon than most modern missiles. It is believed that the attack was \textcolor{blue}{part of the Kremlin's attempts to retake a town, Soledar, and the nearby city of Bakhmut, as a way to grind down the best Ukrainian forces and prevent them from launching counterattacks} elsewhere. Nine people were killed, including a 15-year old girl, Kyrylo Tymoshenko, and \textcolor{red}{Russia denied responsibility for the attack}, claiming it was caused by Ukrainian air defences.
     & 
     \textbf{Summary:} Ukraine's air defense systems have demonstrated their capability in repelling the missile attacks from Russian forces in Kyiv. [1] reported that \textcolor{blue}{Ukrainian air defense forces shot down 37 of "about 40" rockets that entered the city's airspace}, with one person injured. [2] revealed that a commander, who asked to be identified only by the call sign "Hera" for security reasons, stated that \textcolor{blue}{a cruise missile was shot down by a machine gun} from the city's territorial defense mobile group. [3] mentioned that U.S. President Joe Biden had \textcolor{blue}{pledged to deliver one Patriot surface-to-air missile battery system} to Ukraine, which is one of the most advanced U.S. air defense systems and is \textcolor{blue}{capable of intercepting threats such as aircraft and ballistic missiles.}\\
     \hline
    \end{tabular}
    \caption{Table showing summaries for two strategic questions corresponding to a \name{} chapter on Russian missile attacks. The tactically useful and relevant information has been highlighted in \textcolor{blue}{blue}. Tactically useful but irrelevant information has been highlighted in \textcolor{red}{red}.}
    \label{tab:tactical_info_examples}
    \vspace{-2em}
\end{table}

\subsection{\name{} Limitations}
\begin{itemize}
    \item The generation of situation reports within \name{} leverages news articles aggregated from Google News. Nonetheless, the process does not involve a rigorous assessment of the news sources' credibility, nor does it incorporate an explicit verification of the factual accuracy of the claims used in the summary generation model. Considering the extensive exploration of these aspects within computational social science~\cite{lee2023pandemic} and natural language processing~\cite{gong2023fake}, we deemed them beyond the scope of \name{}'s framework in this work.
    \item \name{} enhances the reliability and credibility of the generated situation reports by incorporating citations. However, it does not rigorously ensure the accuracy of these reports in reflecting the content of the source documents.
     In contrast, LLMs have been shown~\cite{mallen2022not, baek2023knowledge} to tend to produce content that may not directly correlate to the source materials. While this is still an active area of research, recent studies~\cite{tian2023fine} have shown that LLMs can be effectively optimized to enhance factual accuracy and attribution capabilities~\cite{gao2023enabling}. Consequently, we propose that future advancements could involve substituting the current summary-generating language model with one that is specifically refined for improved attribution~\cite{gao2023enabling}.
     \item The selection of analysts for the user studies in \name{} was primarily comprised of individuals from military intelligence. This limitation arose due to the challenges encountered in recruiting analysts who could participate without breaching confidentiality agreements. Consequently, our recruitment efforts were confined to analysts within our existing networks. It is important to recognize that our studies may not fully represent the needs and perspectives of analysts in other fields, such as political, corporate, or criminal analysis. Acknowledging this gap, future work can aim to extend and adapt \name{}'s capabilities for generating situation reports, tailoring them to meet the distinct requirements of each of these domains.
\end{itemize}

\section*{Acknowledgement}
We would like to acknowledge the invaluable contributions of Paul Sullivan, who passed away before the publication of this paper. His enduring commitment to knowledge is greatly missed and deeply appreciated. We are grateful to Lisa Ferro and Brad Goodman from MITRE for their valuable comments and help with expert evaluation. This research is based upon work supported by U.S. DARPA AIDA Program No. FA8750-18-2-0014, DARPA KAIROS Program No. FA8750-19-2-1004, DARPA SemaFor Program No. HR001120C0123, DARPA INCAS Program No. HR001121C0165 and DARPA MIPS Program No. HR00112290105. The views and conclusions contained herein are those of the authors and should not be interpreted as necessarily representing the official policies, either expressed or implied, of DARPA, or the U.S. Government. The U.S. Government is authorized to reproduce and distribute reprints for governmental purposes notwithstanding any copyright annotation therein.

%% file: tables/intelligence_analyst_demographic.tex
\begin{table}[!htb]
\centering
\label{tab:participants}
\begin{tabular}{llllll}
\toprule
\textbf{PID} &
  \textbf{Age} &
  \textbf{Education} &
  \textbf{Intelligence Exp.} &
  \textbf{AI Knowledge} &
  \textbf{LLM Usage} \\\midrule
IA1 & 25 - 34 & Bachelor's & 5 - 10 years & 2 - Intermediate & Rarely \\
IA2 & 25 - 34 & High School Diploma & 2 - 5 years & 3 - Proficient & Rarely  \\ 
IA3 & 18 - 24 & High School Diploma & 1 - 2 years & 2 - Intermediate & Daily  \\
IA4 & 25 - 34 & Bachelor's & 2 - 5 years & 2 - Intermediate & Rarely  \\
IA5 & 45 - 54 & Bachelor's & 2 - 5 years & 1 - Basic & Rarely  \\
IA6 & 45 - 54 & Master's & 1 - 2 years & 1  - Basic & Rarely  \\
IA7 & 35 - 44 & High School Diploma & 5 - 10 years & 1  - Basic & Weekly  \\
IA8 & 45 - 54 & Master's & 1 - 2 years & 2 - Intermediate & Rarely  \\
IA9 & 35 - 44 & Bachelor's & 2 - 5 years & 2 - Intermediate & Rarely  \\
IA10 & 25 - 34 & Bachelor & 2 - 5 years & 4 - Advanced & Weekly   \\ 
\bottomrule      
\end{tabular}
\vspace{0.3em}
\caption{Demographic information of study participants along with their experience in intelligence report generation.}
\label{tab:analyst_demographic}
\vspace{-1.5em}
\end{table}

%% file: main.bbl

\begin{thebibliography}{84}


\ifx \showCODEN    \undefined \def \showCODEN     #1{\unskip}     \fi
\ifx \showDOI      \undefined \def \showDOI       #1{#1}\fi
\ifx \showISBNx    \undefined \def \showISBNx     #1{\unskip}     \fi
\ifx \showISBNxiii \undefined \def \showISBNxiii  #1{\unskip}     \fi
\ifx \showISSN     \undefined \def \showISSN      #1{\unskip}     \fi
\ifx \showLCCN     \undefined \def \showLCCN      #1{\unskip}     \fi
\ifx \shownote     \undefined \def \shownote      #1{#1}          \fi
\ifx \showarticletitle \undefined \def \showarticletitle #1{#1}   \fi
\ifx \showURL      \undefined \def \showURL       {\relax}        \fi
\providecommand\bibfield[2]{#2}
\providecommand\bibinfo[2]{#2}
\providecommand\natexlab[1]{#1}
\providecommand\showeprint[2][]{arXiv:#2}

\bibitem[Abdi et~al\mbox{.}(2017)]%
        {abdi2017query}
\bibfield{author}{\bibinfo{person}{Asad Abdi}, \bibinfo{person}{Norisma Idris}, \bibinfo{person}{Rasim~M Alguliyev}, {and} \bibinfo{person}{Ramiz~M Aliguliyev}.} \bibinfo{year}{2017}\natexlab{}.
\newblock \showarticletitle{Query-based multi-documents summarization using linguistic knowledge and content word expansion}.
\newblock \bibinfo{journal}{\emph{Soft Computing}} \bibinfo{volume}{21}, \bibinfo{number}{7} (\bibinfo{year}{2017}), \bibinfo{pages}{1785--1801}.
\newblock


\bibitem[Achiam et~al\mbox{.}(2023)]%
        {achiam2023gpt}
\bibfield{author}{\bibinfo{person}{Josh Achiam}, \bibinfo{person}{Steven Adler}, \bibinfo{person}{Sandhini Agarwal}, \bibinfo{person}{Lama Ahmad}, \bibinfo{person}{Ilge Akkaya}, \bibinfo{person}{Florencia~Leoni Aleman}, \bibinfo{person}{Diogo Almeida}, \bibinfo{person}{Janko Altenschmidt}, \bibinfo{person}{Sam Altman}, \bibinfo{person}{Shyamal Anadkat}, {et~al\mbox{.}}} \bibinfo{year}{2023}\natexlab{}.
\newblock \showarticletitle{Gpt-4 technical report}.
\newblock \bibinfo{journal}{\emph{arXiv preprint arXiv:2303.08774}} (\bibinfo{year}{2023}).
\newblock


\bibitem[Baek et~al\mbox{.}(2023)]%
        {baek2023knowledge}
\bibfield{author}{\bibinfo{person}{Jinheon Baek}, \bibinfo{person}{Soyeong Jeong}, \bibinfo{person}{Minki Kang}, \bibinfo{person}{Jong~C Park}, {and} \bibinfo{person}{Sung~Ju Hwang}.} \bibinfo{year}{2023}\natexlab{}.
\newblock \showarticletitle{Knowledge-Augmented Language Model Verification}.
\newblock \bibinfo{journal}{\emph{arXiv preprint arXiv:2310.12836}} (\bibinfo{year}{2023}).
\newblock


\bibitem[Bai et~al\mbox{.}(2022)]%
        {bai2022training}
\bibfield{author}{\bibinfo{person}{Yuntao Bai}, \bibinfo{person}{Andy Jones}, \bibinfo{person}{Kamal Ndousse}, \bibinfo{person}{Amanda Askell}, \bibinfo{person}{Anna Chen}, \bibinfo{person}{Nova DasSarma}, \bibinfo{person}{Dawn Drain}, \bibinfo{person}{Stanislav Fort}, \bibinfo{person}{Deep Ganguli}, \bibinfo{person}{Tom Henighan}, {et~al\mbox{.}}} \bibinfo{year}{2022}\natexlab{}.
\newblock \showarticletitle{Training a helpful and harmless assistant with reinforcement learning from human feedback}.
\newblock \bibinfo{journal}{\emph{arXiv preprint arXiv:2204.05862}} (\bibinfo{year}{2022}).
\newblock


\bibitem[Bang et~al\mbox{.}(2023)]%
        {bang2023multitask}
\bibfield{author}{\bibinfo{person}{Yejin Bang}, \bibinfo{person}{Samuel Cahyawijaya}, \bibinfo{person}{Nayeon Lee}, \bibinfo{person}{Wenliang Dai}, \bibinfo{person}{Dan Su}, \bibinfo{person}{Bryan Wilie}, \bibinfo{person}{Holy Lovenia}, \bibinfo{person}{Ziwei Ji}, \bibinfo{person}{Tiezheng Yu}, \bibinfo{person}{Willy Chung}, {et~al\mbox{.}}} \bibinfo{year}{2023}\natexlab{}.
\newblock \showarticletitle{A Multitask, Multilingual, Multimodal Evaluation of ChatGPT on Reasoning, Hallucination, and Interactivity}. In \bibinfo{booktitle}{\emph{Proceedings of the 13th International Joint Conference on Natural Language Processing and the 3rd Conference of the Asia-Pacific Chapter of the Association for Computational Linguistics (Volume 1: Long Papers)}}. \bibinfo{pages}{675--718}.
\newblock


\bibitem[Belton and Dhami(2020)]%
        {belton2020cognitive}
\bibfield{author}{\bibinfo{person}{Ian~K Belton} {and} \bibinfo{person}{Mandeep~K Dhami}.} \bibinfo{year}{2020}\natexlab{}.
\newblock \showarticletitle{Cognitive biases and debiasing in intelligence analysis}.
\newblock In \bibinfo{booktitle}{\emph{Routledge Handbook of Bounded Rationality}}. \bibinfo{publisher}{Routledge}, \bibinfo{pages}{548--560}.
\newblock


\bibitem[Bhaskar et~al\mbox{.}(2022)]%
        {bhaskar2022zero}
\bibfield{author}{\bibinfo{person}{Adithya Bhaskar}, \bibinfo{person}{Alexander~R Fabbri}, {and} \bibinfo{person}{Greg Durrett}.} \bibinfo{year}{2022}\natexlab{}.
\newblock \showarticletitle{Zero-Shot Opinion Summarization with GPT-3}.
\newblock \bibinfo{journal}{\emph{arXiv preprint arXiv:2211.15914}} (\bibinfo{year}{2022}).
\newblock


\bibitem[Bohnet et~al\mbox{.}(2022)]%
        {bohnet2022attributed}
\bibfield{author}{\bibinfo{person}{Bernd Bohnet}, \bibinfo{person}{Vinh~Q Tran}, \bibinfo{person}{Pat Verga}, \bibinfo{person}{Roee Aharoni}, \bibinfo{person}{Daniel Andor}, \bibinfo{person}{Livio~Baldini Soares}, \bibinfo{person}{Massimiliano Ciaramita}, \bibinfo{person}{Jacob Eisenstein}, \bibinfo{person}{Kuzman Ganchev}, \bibinfo{person}{Jonathan Herzig}, {et~al\mbox{.}}} \bibinfo{year}{2022}\natexlab{}.
\newblock \showarticletitle{Attributed question answering: Evaluation and modeling for attributed large language models}.
\newblock \bibinfo{journal}{\emph{arXiv preprint arXiv:2212.08037}} (\bibinfo{year}{2022}).
\newblock


\bibitem[Brown et~al\mbox{.}(2020)]%
        {brown2020language}
\bibfield{author}{\bibinfo{person}{Tom Brown}, \bibinfo{person}{Benjamin Mann}, \bibinfo{person}{Nick Ryder}, \bibinfo{person}{Melanie Subbiah}, \bibinfo{person}{Jared~D Kaplan}, \bibinfo{person}{Prafulla Dhariwal}, \bibinfo{person}{Arvind Neelakantan}, \bibinfo{person}{Pranav Shyam}, \bibinfo{person}{Girish Sastry}, \bibinfo{person}{Amanda Askell}, {et~al\mbox{.}}} \bibinfo{year}{2020}\natexlab{}.
\newblock \showarticletitle{Language models are few-shot learners}.
\newblock \bibinfo{journal}{\emph{Advances in neural information processing systems}}  \bibinfo{volume}{33} (\bibinfo{year}{2020}), \bibinfo{pages}{1877--1901}.
\newblock


\bibitem[Cardon et~al\mbox{.}(2023)]%
        {cardon2023challenges}
\bibfield{author}{\bibinfo{person}{Peter Cardon}, \bibinfo{person}{Carolin Fleischmann}, \bibinfo{person}{Jolanta Aritz}, \bibinfo{person}{Minna Logemann}, {and} \bibinfo{person}{Jeanette Heidewald}.} \bibinfo{year}{2023}\natexlab{}.
\newblock \showarticletitle{The Challenges and Opportunities of AI-Assisted Writing: Developing AI Literacy for the AI Age}.
\newblock \bibinfo{journal}{\emph{Business and Professional Communication Quarterly}} (\bibinfo{year}{2023}), \bibinfo{pages}{23294906231176517}.
\newblock


\bibitem[Castonguay(2013)]%
        {biases1}
\bibfield{author}{\bibinfo{person}{Major~J.R.P. Castonguay}.} \bibinfo{year}{2013}\natexlab{}.
\newblock \bibinfo{title}{INTELLIGENCE ANALYSIS: FLAWED WITH BIASES}.
\newblock
\newblock
\newblock
\shownote{CANADIAN FORCES COLLEGE. https://www.cfc.forces.gc.ca/259/290/301/305/castonguay.pdf}.


\bibitem[Chakravarti and Sil(2021)]%
        {chakravarti2021towards}
\bibfield{author}{\bibinfo{person}{Rishav Chakravarti} {and} \bibinfo{person}{Avirup Sil}.} \bibinfo{year}{2021}\natexlab{}.
\newblock \showarticletitle{Towards confident machine reading comprehension}.
\newblock \bibinfo{journal}{\emph{arXiv preprint arXiv:2101.07942}} (\bibinfo{year}{2021}).
\newblock


\bibitem[Chen et~al\mbox{.}(2021)]%
        {chen2021evaluating}
\bibfield{author}{\bibinfo{person}{Mark Chen}, \bibinfo{person}{Jerry Tworek}, \bibinfo{person}{Heewoo Jun}, \bibinfo{person}{Qiming Yuan}, \bibinfo{person}{Henrique Ponde de~Oliveira Pinto}, \bibinfo{person}{Jared Kaplan}, \bibinfo{person}{Harri Edwards}, \bibinfo{person}{Yuri Burda}, \bibinfo{person}{Nicholas Joseph}, \bibinfo{person}{Greg Brockman}, {et~al\mbox{.}}} \bibinfo{year}{2021}\natexlab{}.
\newblock \showarticletitle{Evaluating large language models trained on code}.
\newblock \bibinfo{journal}{\emph{arXiv preprint arXiv:2107.03374}} (\bibinfo{year}{2021}).
\newblock


\bibitem[Chowdhery et~al\mbox{.}(2022)]%
        {chowdhery2022palm}
\bibfield{author}{\bibinfo{person}{Aakanksha Chowdhery}, \bibinfo{person}{Sharan Narang}, \bibinfo{person}{Jacob Devlin}, \bibinfo{person}{Maarten Bosma}, \bibinfo{person}{Gaurav Mishra}, \bibinfo{person}{Adam Roberts}, \bibinfo{person}{Paul Barham}, \bibinfo{person}{Hyung~Won Chung}, \bibinfo{person}{Charles Sutton}, \bibinfo{person}{Sebastian Gehrmann}, {et~al\mbox{.}}} \bibinfo{year}{2022}\natexlab{}.
\newblock \showarticletitle{Palm: Scaling language modeling with pathways}.
\newblock \bibinfo{journal}{\emph{arXiv preprint arXiv:2204.02311}} (\bibinfo{year}{2022}).
\newblock


\bibitem[Council et~al\mbox{.}(2011)]%
        {national2011intelligence}
\bibfield{author}{\bibinfo{person}{National~Research Council} {et~al\mbox{.}}} \bibinfo{year}{2011}\natexlab{}.
\newblock \bibinfo{booktitle}{\emph{Intelligence analysis for tomorrow: Advances from the behavioral and social sciences}}.
\newblock \bibinfo{publisher}{National Academies Press}.
\newblock


\bibitem[Doggette et~al\mbox{.}(2020)]%
        {doggette2020information}
\bibfield{author}{\bibinfo{person}{Brandon~D Doggette}, \bibinfo{person}{US~Army Command}, {and} \bibinfo{person}{General~Staff College}.} \bibinfo{year}{2020}\natexlab{}.
\newblock \emph{\bibinfo{title}{Information overload: impacts on brigade combat team S-2 current operations intelligence analysts}}.
\newblock \bibinfo{thesistype}{Ph.\,D. Dissertation}. \bibinfo{school}{Fort Leavenworth, KS: US Army Command and General Staff College}.
\newblock


\bibitem[Du et~al\mbox{.}(2017)]%
        {du-etal-2017-learning}
\bibfield{author}{\bibinfo{person}{Xinya Du}, \bibinfo{person}{Junru Shao}, {and} \bibinfo{person}{Claire Cardie}.} \bibinfo{year}{2017}\natexlab{}.
\newblock \showarticletitle{Learning to Ask: Neural Question Generation for Reading Comprehension}. In \bibinfo{booktitle}{\emph{Proceedings of the 55th Annual Meeting of the Association for Computational Linguistics (Volume 1: Long Papers)}}. \bibinfo{publisher}{Association for Computational Linguistics}, \bibinfo{address}{Vancouver, Canada}, \bibinfo{pages}{1342--1352}.
\newblock
\urldef\tempurl%
\url{https://doi.org/10.18653/v1/P17-1123}
\showDOI{\tempurl}


\bibitem[Du et~al\mbox{.}(2022)]%
        {du2022resin}
\bibfield{author}{\bibinfo{person}{Xinya Du}, \bibinfo{person}{Zixuan Zhang}, \bibinfo{person}{Sha Li}, \bibinfo{person}{Pengfei Yu}, \bibinfo{person}{Hongwei Wang}, \bibinfo{person}{Tuan Lai}, \bibinfo{person}{Xudong Lin}, \bibinfo{person}{Ziqi Wang}, \bibinfo{person}{Iris Liu}, \bibinfo{person}{Ben Zhou}, {et~al\mbox{.}}} \bibinfo{year}{2022}\natexlab{}.
\newblock \showarticletitle{RESIN-11: Schema-guided event prediction for 11 newsworthy scenarios}. In \bibinfo{booktitle}{\emph{Proceedings of the 2022 Conference of the North American Chapter of the Association for Computational Linguistics: Human Language Technologies: System Demonstrations}}. \bibinfo{pages}{54--63}.
\newblock


\bibitem[Fabbri et~al\mbox{.}(2019)]%
        {fabbri-etal-2019-multi}
\bibfield{author}{\bibinfo{person}{Alexander Fabbri}, \bibinfo{person}{Irene Li}, \bibinfo{person}{Tianwei She}, \bibinfo{person}{Suyi Li}, {and} \bibinfo{person}{Dragomir Radev}.} \bibinfo{year}{2019}\natexlab{}.
\newblock \showarticletitle{Multi-News: A Large-Scale Multi-Document Summarization Dataset and Abstractive Hierarchical Model}. In \bibinfo{booktitle}{\emph{Proceedings of the 57th Annual Meeting of the Association for Computational Linguistics}}. \bibinfo{publisher}{Association for Computational Linguistics}, \bibinfo{address}{Florence, Italy}, \bibinfo{pages}{1074--1084}.
\newblock
\urldef\tempurl%
\url{https://doi.org/10.18653/v1/P19-1102}
\showDOI{\tempurl}


\bibitem[Gao et~al\mbox{.}(2023a)]%
        {gao2023rarr}
\bibfield{author}{\bibinfo{person}{Luyu Gao}, \bibinfo{person}{Zhuyun Dai}, \bibinfo{person}{Panupong Pasupat}, \bibinfo{person}{Anthony Chen}, \bibinfo{person}{Arun~Tejasvi Chaganty}, \bibinfo{person}{Yicheng Fan}, \bibinfo{person}{Vincent Zhao}, \bibinfo{person}{Ni Lao}, \bibinfo{person}{Hongrae Lee}, \bibinfo{person}{Da-Cheng Juan}, {et~al\mbox{.}}} \bibinfo{year}{2023}\natexlab{a}.
\newblock \showarticletitle{RARR: Researching and Revising What Language Models Say, Using Language Models}. In \bibinfo{booktitle}{\emph{Proceedings of the 61st Annual Meeting of the Association for Computational Linguistics (Volume 1: Long Papers)}}. \bibinfo{pages}{16477--16508}.
\newblock


\bibitem[Gao et~al\mbox{.}(2023b)]%
        {gao2023enabling}
\bibfield{author}{\bibinfo{person}{Tianyu Gao}, \bibinfo{person}{Howard Yen}, \bibinfo{person}{Jiatong Yu}, {and} \bibinfo{person}{Danqi Chen}.} \bibinfo{year}{2023}\natexlab{b}.
\newblock \showarticletitle{Enabling Large Language Models to Generate Text with Citations}. In \bibinfo{booktitle}{\emph{Proceedings of the 2023 Conference on Empirical Methods in Natural Language Processing}}. \bibinfo{pages}{6465--6488}.
\newblock


\bibitem[Gao et~al\mbox{.}(2023c)]%
        {gao-etal-2023-enabling}
\bibfield{author}{\bibinfo{person}{Tianyu Gao}, \bibinfo{person}{Howard Yen}, \bibinfo{person}{Jiatong Yu}, {and} \bibinfo{person}{Danqi Chen}.} \bibinfo{year}{2023}\natexlab{c}.
\newblock \showarticletitle{Enabling Large Language Models to Generate Text with Citations}. In \bibinfo{booktitle}{\emph{Proceedings of the 2023 Conference on Empirical Methods in Natural Language Processing}}, \bibfield{editor}{\bibinfo{person}{Houda Bouamor}, \bibinfo{person}{Juan Pino}, {and} \bibinfo{person}{Kalika Bali}} (Eds.). \bibinfo{publisher}{Association for Computational Linguistics}, \bibinfo{address}{Singapore}, \bibinfo{pages}{6465--6488}.
\newblock
\urldef\tempurl%
\url{https://aclanthology.org/2023.emnlp-main.398}
\showURL{%
\tempurl}


\bibitem[Garg et~al\mbox{.}(2020)]%
        {Garg_2020}
\bibfield{author}{\bibinfo{person}{Siddhant Garg}, \bibinfo{person}{Thuy Vu}, {and} \bibinfo{person}{Alessandro Moschitti}.} \bibinfo{year}{2020}\natexlab{}.
\newblock \showarticletitle{TANDA: Transfer and Adapt Pre-Trained Transformer Models for Answer Sentence Selection}.
\newblock \bibinfo{journal}{\emph{Proceedings of the AAAI Conference on Artificial Intelligence}} \bibinfo{volume}{34}, \bibinfo{number}{05} (\bibinfo{date}{Apr} \bibinfo{year}{2020}), \bibinfo{pages}{7780–7788}.
\newblock
\showISSN{2159-5399}
\urldef\tempurl%
\url{https://doi.org/10.1609/aaai.v34i05.6282}
\showDOI{\tempurl}


\bibitem[Gero et~al\mbox{.}(2022)]%
        {gero2022sparks}
\bibfield{author}{\bibinfo{person}{Katy~Ilonka Gero}, \bibinfo{person}{Vivian Liu}, {and} \bibinfo{person}{Lydia Chilton}.} \bibinfo{year}{2022}\natexlab{}.
\newblock \showarticletitle{Sparks: Inspiration for science writing using language models}. In \bibinfo{booktitle}{\emph{Designing interactive systems conference}}. \bibinfo{pages}{1002--1019}.
\newblock


\bibitem[Golson and Ferraro(2018)]%
        {golson2018resist}
\bibfield{author}{\bibinfo{person}{Preston Golson} {and} \bibinfo{person}{Matthew~F Ferraro}.} \bibinfo{year}{2018}\natexlab{}.
\newblock \showarticletitle{To Resist Disinformation, Learn to Think Like an Intelligence Analyst}.
\newblock \bibinfo{journal}{\emph{CIA Studies in Intelligence}} \bibinfo{volume}{62}, \bibinfo{number}{1} (\bibinfo{year}{2018}), \bibinfo{pages}{31--32}.
\newblock


\bibitem[Gong et~al\mbox{.}(2023)]%
        {gong2023fake}
\bibfield{author}{\bibinfo{person}{Shuzhi Gong}, \bibinfo{person}{Richard~O Sinnott}, \bibinfo{person}{Jianzhong Qi}, {and} \bibinfo{person}{Cecile Paris}.} \bibinfo{year}{2023}\natexlab{}.
\newblock \showarticletitle{Fake news detection through graph-based neural networks: A survey}.
\newblock \bibinfo{journal}{\emph{arXiv preprint arXiv:2307.12639}} (\bibinfo{year}{2023}).
\newblock


\bibitem[Goyal et~al\mbox{.}(2022)]%
        {goyal2022news}
\bibfield{author}{\bibinfo{person}{Tanya Goyal}, \bibinfo{person}{Junyi~Jessy Li}, {and} \bibinfo{person}{Greg Durrett}.} \bibinfo{year}{2022}\natexlab{}.
\newblock \showarticletitle{News Summarization and Evaluation in the Era of GPT-3}.
\newblock \bibinfo{journal}{\emph{arXiv preprint arXiv:2209.12356}} (\bibinfo{year}{2022}).
\newblock


\bibitem[Gu et~al\mbox{.}(2020)]%
        {headline2020}
\bibfield{author}{\bibinfo{person}{Xiaotao Gu}, \bibinfo{person}{Yuning Mao}, \bibinfo{person}{Jiawei Han}, \bibinfo{person}{Jialu Liu}, \bibinfo{person}{Hongkun Yu}, \bibinfo{person}{You Wu}, \bibinfo{person}{Cong Yu}, \bibinfo{person}{Daniel Finnie}, \bibinfo{person}{Jiaqi Zhai}, {and} \bibinfo{person}{Nicholas Zukoski}.} \bibinfo{year}{2020}\natexlab{}.
\newblock \showarticletitle{{Generating Representative Headlines for News Stories}}. In \bibinfo{booktitle}{\emph{Proc. of the the Web Conf. 2020}}.
\newblock


\bibitem[Holtzman et~al\mbox{.}(2019)]%
        {holtzman2019curious}
\bibfield{author}{\bibinfo{person}{Ari Holtzman}, \bibinfo{person}{Jan Buys}, \bibinfo{person}{Li Du}, \bibinfo{person}{Maxwell Forbes}, {and} \bibinfo{person}{Yejin Choi}.} \bibinfo{year}{2019}\natexlab{}.
\newblock \showarticletitle{The Curious Case of Neural Text Degeneration}. In \bibinfo{booktitle}{\emph{International Conference on Learning Representations}}.
\newblock


\bibitem[Honovich et~al\mbox{.}(2022)]%
        {honovich-etal-2022-true-evaluating}
\bibfield{author}{\bibinfo{person}{Or Honovich}, \bibinfo{person}{Roee Aharoni}, \bibinfo{person}{Jonathan Herzig}, \bibinfo{person}{Hagai Taitelbaum}, \bibinfo{person}{Doron Kukliansy}, \bibinfo{person}{Vered Cohen}, \bibinfo{person}{Thomas Scialom}, \bibinfo{person}{Idan Szpektor}, \bibinfo{person}{Avinatan Hassidim}, {and} \bibinfo{person}{Yossi Matias}.} \bibinfo{year}{2022}\natexlab{}.
\newblock \showarticletitle{{TRUE}: Re-evaluating Factual Consistency Evaluation}. In \bibinfo{booktitle}{\emph{Proceedings of the 2022 Conference of the North American Chapter of the Association for Computational Linguistics: Human Language Technologies}}, \bibfield{editor}{\bibinfo{person}{Marine Carpuat}, \bibinfo{person}{Marie-Catherine de~Marneffe}, {and} \bibinfo{person}{Ivan~Vladimir Meza~Ruiz}} (Eds.). \bibinfo{publisher}{Association for Computational Linguistics}, \bibinfo{address}{Seattle, United States}, \bibinfo{pages}{3905--3920}.
\newblock
\urldef\tempurl%
\url{https://doi.org/10.18653/v1/2022.naacl-main.287}
\showDOI{\tempurl}


\bibitem[Jain and Dubes(1988)]%
        {jain1988algorithms}
\bibfield{author}{\bibinfo{person}{Anil~K Jain} {and} \bibinfo{person}{Richard~C Dubes}.} \bibinfo{year}{1988}\natexlab{}.
\newblock \showarticletitle{Algorithms for clustering data}.
\newblock  (\bibinfo{year}{1988}).
\newblock


\bibitem[Ji et~al\mbox{.}({[n.\,d.]})]%
        {ji2022survey}
\bibfield{author}{\bibinfo{person}{Ziwei Ji}, \bibinfo{person}{Nayeon Lee}, \bibinfo{person}{Rita Frieske}, \bibinfo{person}{Tiezheng Yu}, \bibinfo{person}{Dan Su}, \bibinfo{person}{Yan Xu}, \bibinfo{person}{Etsuko Ishii}, \bibinfo{person}{Yejin Bang}, \bibinfo{person}{Andrea Madotto}, {and} \bibinfo{person}{Pascale Fung}.} \bibinfo{year}{[n.\,d.]}\natexlab{}.
\newblock \showarticletitle{Survey of hallucination in natural language generation}.
\newblock \bibinfo{journal}{\emph{Comput. Surveys}} (\bibinfo{year}{[n.\,d.]}).
\newblock


\bibitem[Ji et~al\mbox{.}(2023)]%
        {ji2023survey}
\bibfield{author}{\bibinfo{person}{Ziwei Ji}, \bibinfo{person}{Nayeon Lee}, \bibinfo{person}{Rita Frieske}, \bibinfo{person}{Tiezheng Yu}, \bibinfo{person}{Dan Su}, \bibinfo{person}{Yan Xu}, \bibinfo{person}{Etsuko Ishii}, \bibinfo{person}{Ye~Jin Bang}, \bibinfo{person}{Andrea Madotto}, {and} \bibinfo{person}{Pascale Fung}.} \bibinfo{year}{2023}\natexlab{}.
\newblock \showarticletitle{Survey of hallucination in natural language generation}.
\newblock \bibinfo{journal}{\emph{Comput. Surveys}} \bibinfo{volume}{55}, \bibinfo{number}{12} (\bibinfo{year}{2023}), \bibinfo{pages}{1--38}.
\newblock


\bibitem[Kwiatkowski et~al\mbox{.}(2019)]%
        {kwiatkowski2019natural}
\bibfield{author}{\bibinfo{person}{Tom Kwiatkowski}, \bibinfo{person}{Jennimaria Palomaki}, \bibinfo{person}{Olivia Redfield}, \bibinfo{person}{Michael Collins}, \bibinfo{person}{Ankur Parikh}, \bibinfo{person}{Chris Alberti}, \bibinfo{person}{Danielle Epstein}, \bibinfo{person}{Illia Polosukhin}, \bibinfo{person}{Jacob Devlin}, \bibinfo{person}{Kenton Lee}, {et~al\mbox{.}}} \bibinfo{year}{2019}\natexlab{}.
\newblock \showarticletitle{Natural Questions: A Benchmark for Question Answering Research}.
\newblock \bibinfo{journal}{\emph{Transactions of the Association for Computational Linguistics}}  \bibinfo{volume}{7} (\bibinfo{year}{2019}), \bibinfo{pages}{452--466}.
\newblock


\bibitem[Lee et~al\mbox{.}(2023)]%
        {lee2023pandemic}
\bibfield{author}{\bibinfo{person}{Edmund~WJ Lee}, \bibinfo{person}{Huanyu Bao}, \bibinfo{person}{Yixi Wang}, {and} \bibinfo{person}{Yi~Torng Lim}.} \bibinfo{year}{2023}\natexlab{}.
\newblock \showarticletitle{From pandemic to Plandemic: Examining the amplification and attenuation of COVID-19 misinformation on social media}.
\newblock \bibinfo{journal}{\emph{Social Science \& Medicine}}  \bibinfo{volume}{328} (\bibinfo{year}{2023}), \bibinfo{pages}{115979}.
\newblock


\bibitem[Lee et~al\mbox{.}(2022)]%
        {lee2022coauthor}
\bibfield{author}{\bibinfo{person}{Mina Lee}, \bibinfo{person}{Percy Liang}, {and} \bibinfo{person}{Qian Yang}.} \bibinfo{year}{2022}\natexlab{}.
\newblock \showarticletitle{Coauthor: Designing a human-ai collaborative writing dataset for exploring language model capabilities}. In \bibinfo{booktitle}{\emph{Proceedings of the 2022 CHI conference on human factors in computing systems}}. \bibinfo{pages}{1--19}.
\newblock


\bibitem[Lewis et~al\mbox{.}(2020)]%
        {lewis2020bart}
\bibfield{author}{\bibinfo{person}{Mike Lewis}, \bibinfo{person}{Yinhan Liu}, \bibinfo{person}{Naman Goyal}, \bibinfo{person}{Marjan Ghazvininejad}, \bibinfo{person}{Abdelrahman Mohamed}, \bibinfo{person}{Omer Levy}, \bibinfo{person}{Veselin Stoyanov}, {and} \bibinfo{person}{Luke Zettlemoyer}.} \bibinfo{year}{2020}\natexlab{}.
\newblock \showarticletitle{BART: Denoising Sequence-to-Sequence Pre-training for Natural Language Generation, Translation, and Comprehension}. In \bibinfo{booktitle}{\emph{Proceedings of the 58th Annual Meeting of the Association for Computational Linguistics}}. \bibinfo{pages}{7871--7880}.
\newblock


\bibitem[Li et~al\mbox{.}(2014)]%
        {li2014improving}
\bibfield{author}{\bibinfo{person}{Chen Li}, \bibinfo{person}{Yang Liu}, \bibinfo{person}{Fei Liu}, \bibinfo{person}{Lin Zhao}, {and} \bibinfo{person}{Fuliang Weng}.} \bibinfo{year}{2014}\natexlab{}.
\newblock \showarticletitle{Improving multi-documents summarization by sentence compression based on expanded constituent parse trees}. In \bibinfo{booktitle}{\emph{Proceedings of the 2014 Conference on Empirical Methods in Natural Language Processing (EMNLP)}}. \bibinfo{pages}{691--701}.
\newblock


\bibitem[Li et~al\mbox{.}(2023a)]%
        {li2023halueval}
\bibfield{author}{\bibinfo{person}{Junyi Li}, \bibinfo{person}{Xiaoxue Cheng}, \bibinfo{person}{Wayne~Xin Zhao}, \bibinfo{person}{Jian-Yun Nie}, {and} \bibinfo{person}{Ji-Rong Wen}.} \bibinfo{year}{2023}\natexlab{a}.
\newblock \showarticletitle{Halueval: A large-scale hallucination evaluation benchmark for large language models}. In \bibinfo{booktitle}{\emph{Proceedings of the 2023 Conference on Empirical Methods in Natural Language Processing}}. \bibinfo{pages}{6449--6464}.
\newblock


\bibitem[Li et~al\mbox{.}(2021)]%
        {li2021future}
\bibfield{author}{\bibinfo{person}{Manling Li}, \bibinfo{person}{Sha Li}, \bibinfo{person}{Zhenhailong Wang}, \bibinfo{person}{Lifu Huang}, \bibinfo{person}{Kyunghyun Cho}, \bibinfo{person}{Heng Ji}, \bibinfo{person}{Jiawei Han}, {and} \bibinfo{person}{Clare Voss}.} \bibinfo{year}{2021}\natexlab{}.
\newblock \showarticletitle{The Future is not One-dimensional: Complex Event Schema Induction by Graph Modeling for Event Prediction}. In \bibinfo{booktitle}{\emph{Proceedings of the 2021 Conference on Empirical Methods in Natural Language Processing}}. \bibinfo{pages}{5203--5215}.
\newblock


\bibitem[Li et~al\mbox{.}(2020)]%
        {li2020gaia}
\bibfield{author}{\bibinfo{person}{Manling Li}, \bibinfo{person}{Alireza Zareian}, \bibinfo{person}{Ying Lin}, \bibinfo{person}{Xiaoman Pan}, \bibinfo{person}{Spencer Whitehead}, \bibinfo{person}{Brian Chen}, \bibinfo{person}{Bo Wu}, \bibinfo{person}{Heng Ji}, \bibinfo{person}{Shih-Fu Chang}, \bibinfo{person}{Clare Voss}, {et~al\mbox{.}}} \bibinfo{year}{2020}\natexlab{}.
\newblock \showarticletitle{Gaia: A fine-grained multimedia knowledge extraction system}. In \bibinfo{booktitle}{\emph{Proceedings of the 58th Annual Meeting of the Association for Computational Linguistics: System Demonstrations}}. \bibinfo{pages}{77--86}.
\newblock


\bibitem[Li et~al\mbox{.}(2023b)]%
        {li2023open}
\bibfield{author}{\bibinfo{person}{Sha Li}, \bibinfo{person}{Ruining Zhao}, \bibinfo{person}{Manling Li}, \bibinfo{person}{Heng Ji}, \bibinfo{person}{Chris Callison-Burch}, {and} \bibinfo{person}{Jiawei Han}.} \bibinfo{year}{2023}\natexlab{b}.
\newblock \showarticletitle{Open-Domain Hierarchical Event Schema Induction by Incremental Prompting and Verification}. In \bibinfo{booktitle}{\emph{Proceedings of the 61st Annual Meeting of the Association for Computational Linguistics (Volume 1: Long Papers)}}. \bibinfo{pages}{5677--5697}.
\newblock


\bibitem[Lin(2004)]%
        {lin2004rouge}
\bibfield{author}{\bibinfo{person}{Chin-Yew Lin}.} \bibinfo{year}{2004}\natexlab{}.
\newblock \showarticletitle{Rouge: A package for automatic evaluation of summaries}. In \bibinfo{booktitle}{\emph{Text summarization branches out}}. \bibinfo{pages}{74--81}.
\newblock


\bibitem[Liu et~al\mbox{.}(2022)]%
        {liu2022brio}
\bibfield{author}{\bibinfo{person}{Yixin Liu}, \bibinfo{person}{Pengfei Liu}, \bibinfo{person}{Dragomir Radev}, {and} \bibinfo{person}{Graham Neubig}.} \bibinfo{year}{2022}\natexlab{}.
\newblock \showarticletitle{BRIO: Bringing Order to Abstractive Summarization}. In \bibinfo{booktitle}{\emph{Proceedings of the 60th Annual Meeting of the Association for Computational Linguistics (Volume 1: Long Papers)}}. \bibinfo{pages}{2890--2903}.
\newblock


\bibitem[Liu et~al\mbox{.}(2019)]%
        {liu2019roberta}
\bibfield{author}{\bibinfo{person}{Yinhan Liu}, \bibinfo{person}{Myle Ott}, \bibinfo{person}{Naman Goyal}, \bibinfo{person}{Jingfei Du}, \bibinfo{person}{Mandar Joshi}, \bibinfo{person}{Danqi Chen}, \bibinfo{person}{Omer Levy}, \bibinfo{person}{Mike Lewis}, \bibinfo{person}{Luke Zettlemoyer}, {and} \bibinfo{person}{Veselin Stoyanov}.} \bibinfo{year}{2019}\natexlab{}.
\newblock \showarticletitle{Roberta: A robustly optimized bert pretraining approach}.
\newblock \bibinfo{journal}{\emph{arXiv preprint arXiv:1907.11692}} (\bibinfo{year}{2019}).
\newblock


\bibitem[Ma et~al\mbox{.}(2023)]%
        {ma2023structured}
\bibfield{author}{\bibinfo{person}{Yunshan Ma}, \bibinfo{person}{Chenchen Ye}, \bibinfo{person}{Zijian Wu}, \bibinfo{person}{Xiang Wang}, \bibinfo{person}{Yixin Cao}, \bibinfo{person}{Liang Pang}, {and} \bibinfo{person}{Tat-Seng Chua}.} \bibinfo{year}{2023}\natexlab{}.
\newblock \showarticletitle{Structured, Complex and Time-complete Temporal Event Forecasting}.
\newblock \bibinfo{journal}{\emph{arXiv preprint arXiv:2312.01052}} (\bibinfo{year}{2023}).
\newblock


\bibitem[Mallen et~al\mbox{.}(2022)]%
        {mallen2022not}
\bibfield{author}{\bibinfo{person}{Alex Mallen}, \bibinfo{person}{Akari Asai}, \bibinfo{person}{Victor Zhong}, \bibinfo{person}{Rajarshi Das}, \bibinfo{person}{Hannaneh Hajishirzi}, {and} \bibinfo{person}{Daniel Khashabi}.} \bibinfo{year}{2022}\natexlab{}.
\newblock \showarticletitle{When not to trust language models: Investigating effectiveness and limitations of parametric and non-parametric memories}.
\newblock \bibinfo{journal}{\emph{arXiv preprint arXiv:2212.10511}} (\bibinfo{year}{2022}).
\newblock


\bibitem[Maynez et~al\mbox{.}(2020)]%
        {maynez2020faithfulness}
\bibfield{author}{\bibinfo{person}{Joshua Maynez}, \bibinfo{person}{Shashi Narayan}, \bibinfo{person}{Bernd Bohnet}, {and} \bibinfo{person}{Ryan McDonald}.} \bibinfo{year}{2020}\natexlab{}.
\newblock \showarticletitle{On Faithfulness and Factuality in Abstractive Summarization}. In \bibinfo{booktitle}{\emph{Proceedings of the 58th Annual Meeting of the Association for Computational Linguistics}}. \bibinfo{pages}{1906--1919}.
\newblock


\bibitem[Mirowski et~al\mbox{.}(2023)]%
        {mirowski2023co}
\bibfield{author}{\bibinfo{person}{Piotr Mirowski}, \bibinfo{person}{Kory~W Mathewson}, \bibinfo{person}{Jaylen Pittman}, {and} \bibinfo{person}{Richard Evans}.} \bibinfo{year}{2023}\natexlab{}.
\newblock \showarticletitle{Co-Writing Screenplays and Theatre Scripts with Language Models: Evaluation by Industry Professionals}. In \bibinfo{booktitle}{\emph{Proceedings of the 2023 CHI Conference on Human Factors in Computing Systems}}. \bibinfo{pages}{1--34}.
\newblock


\bibitem[Monzer et~al\mbox{.}(2020)]%
        {monzer2020user}
\bibfield{author}{\bibinfo{person}{Cristina Monzer}, \bibinfo{person}{Judith Moeller}, \bibinfo{person}{Natali Helberger}, {and} \bibinfo{person}{Sarah Eskens}.} \bibinfo{year}{2020}\natexlab{}.
\newblock \showarticletitle{User perspectives on the news personalisation process: Agency, trust and utility as building blocks}.
\newblock \bibinfo{journal}{\emph{Digital Journalism}} \bibinfo{volume}{8}, \bibinfo{number}{9} (\bibinfo{year}{2020}), \bibinfo{pages}{1142--1162}.
\newblock


\bibitem[Murakhovs{'}ka et~al\mbox{.}(2022)]%
        {murakhovska-etal-2022-mixqg}
\bibfield{author}{\bibinfo{person}{Lidiya Murakhovs{'}ka}, \bibinfo{person}{Chien-Sheng Wu}, \bibinfo{person}{Philippe Laban}, \bibinfo{person}{Tong Niu}, \bibinfo{person}{Wenhao Liu}, {and} \bibinfo{person}{Caiming Xiong}.} \bibinfo{year}{2022}\natexlab{}.
\newblock \showarticletitle{{M}ix{QG}: Neural Question Generation with Mixed Answer Types}. In \bibinfo{booktitle}{\emph{Findings of the Association for Computational Linguistics: NAACL 2022}}. \bibinfo{publisher}{Association for Computational Linguistics}, \bibinfo{address}{Seattle, United States}, \bibinfo{pages}{1486--1497}.
\newblock
\urldef\tempurl%
\url{https://doi.org/10.18653/v1/2022.findings-naacl.111}
\showDOI{\tempurl}


\bibitem[National Academies~of Sciences et~al\mbox{.}(2019)]%
        {national2019decadal}
\bibfield{author}{\bibinfo{person}{Engineering National Academies~of Sciences}, \bibinfo{person}{Medicine}, {et~al\mbox{.}}} \bibinfo{year}{2019}\natexlab{}.
\newblock \showarticletitle{A decadal survey of the social and behavioral sciences: A research agenda for advancing intelligence analysis}.
\newblock  (\bibinfo{year}{2019}).
\newblock


\bibitem[{OpenAI}(2021)]%
        {openai2021gpt35}
\bibfield{author}{\bibinfo{person}{{OpenAI}}.} \bibinfo{year}{2021}\natexlab{}.
\newblock \bibinfo{booktitle}{\emph{GPT-3.5}}.
\newblock
\urldef\tempurl%
\url{https://openai.com/research/}
\showURL{%
\tempurl}
\newblock
\shownote{Computer software}.


\bibitem[Organization et~al\mbox{.}(2020)]%
        {world2020coronavirus}
\bibfield{author}{\bibinfo{person}{World~Health Organization} {et~al\mbox{.}}} \bibinfo{year}{2020}\natexlab{}.
\newblock \showarticletitle{Coronavirus disease 2019 (COVID-19): situation report, 73}.
\newblock  (\bibinfo{year}{2020}).
\newblock


\bibitem[Organization et~al\mbox{.}(2022)]%
        {world2022emergency}
\bibfield{author}{\bibinfo{person}{World~Health Organization} {et~al\mbox{.}}} \bibinfo{year}{2022}\natexlab{}.
\newblock \bibinfo{booktitle}{\emph{Emergency in Ukraine: external situation report\# 13, published 26 May 2022: reporting period: 19--25 May 2022}}.
\newblock \bibinfo{type}{{T}echnical {R}eport}. \bibinfo{institution}{World Health Organization. Regional Office for Europe}.
\newblock


\bibitem[Papineni et~al\mbox{.}(2002)]%
        {papineni2002bleu}
\bibfield{author}{\bibinfo{person}{Kishore Papineni}, \bibinfo{person}{Salim Roukos}, \bibinfo{person}{Todd Ward}, {and} \bibinfo{person}{Wei-Jing Zhu}.} \bibinfo{year}{2002}\natexlab{}.
\newblock \showarticletitle{Bleu: a method for automatic evaluation of machine translation}. In \bibinfo{booktitle}{\emph{Proceedings of the 40th annual meeting of the Association for Computational Linguistics}}. \bibinfo{pages}{311--318}.
\newblock


\bibitem[Rajpurkar et~al\mbox{.}(2018)]%
        {rajpurkar2018know}
\bibfield{author}{\bibinfo{person}{Pranav Rajpurkar}, \bibinfo{person}{Robin Jia}, {and} \bibinfo{person}{Percy Liang}.} \bibinfo{year}{2018}\natexlab{}.
\newblock \showarticletitle{Know What You Don’t Know: Unanswerable Questions for SQuAD}. In \bibinfo{booktitle}{\emph{Proceedings of the 56th Annual Meeting of the Association for Computational Linguistics (Volume 2: Short Papers)}}. \bibinfo{pages}{784--789}.
\newblock


\bibitem[Reddy et~al\mbox{.}(2022a)]%
        {reddy2022zero}
\bibfield{author}{\bibinfo{person}{Revanth~Gangi Reddy}, \bibinfo{person}{Sai~Chetan Chinthakindi}, \bibinfo{person}{Yi~R Fung}, \bibinfo{person}{Kevin Small}, {and} \bibinfo{person}{Heng Ji}.} \bibinfo{year}{2022}\natexlab{a}.
\newblock \showarticletitle{A Zero-Shot Claim Detection Framework using Question Answering}. In \bibinfo{booktitle}{\emph{Proceedings of the 29th International Conference on Computational Linguistics}}. \bibinfo{pages}{6927--6933}.
\newblock


\bibitem[Reddy et~al\mbox{.}(2022b)]%
        {reddy2022newsclaims}
\bibfield{author}{\bibinfo{person}{Revanth~Gangi Reddy}, \bibinfo{person}{Sai~Chetan Chinthakindi}, \bibinfo{person}{Zhenhailong Wang}, \bibinfo{person}{Yi Fung}, \bibinfo{person}{Kathryn Conger}, \bibinfo{person}{Ahmed Elsayed}, \bibinfo{person}{Martha Palmer}, \bibinfo{person}{Preslav Nakov}, \bibinfo{person}{Eduard Hovy}, \bibinfo{person}{Kevin Small}, {et~al\mbox{.}}} \bibinfo{year}{2022}\natexlab{b}.
\newblock \showarticletitle{Newsclaims: A new benchmark for claim detection from news with attribute knowledge}. In \bibinfo{booktitle}{\emph{Proceedings of the 2022 Conference on Empirical Methods in Natural Language Processing}}. \bibinfo{pages}{6002--6018}.
\newblock


\bibitem[Reddy et~al\mbox{.}(2022c)]%
        {reddy2022sumren}
\bibfield{author}{\bibinfo{person}{Revanth~Gangi Reddy}, \bibinfo{person}{Heba Elfardy}, \bibinfo{person}{Hou~Pong Chan}, \bibinfo{person}{Kevin Small}, {and} \bibinfo{person}{Heng Ji}.} \bibinfo{year}{2022}\natexlab{c}.
\newblock \showarticletitle{SumREN: Summarizing Reported Speech about Events in News}.
\newblock \bibinfo{journal}{\emph{arXiv preprint arXiv:2212.01146}} (\bibinfo{year}{2022}).
\newblock


\bibitem[Reddy et~al\mbox{.}(2020)]%
        {reddy2020answer}
\bibfield{author}{\bibinfo{person}{Revanth~Gangi Reddy}, \bibinfo{person}{Md~Arafat Sultan}, \bibinfo{person}{Efsun~Sarioglu Kayi}, \bibinfo{person}{Rong Zhang}, \bibinfo{person}{Vittorio Castelli}, {and} \bibinfo{person}{Avirup Sil}.} \bibinfo{year}{2020}\natexlab{}.
\newblock \showarticletitle{Answer Span Correction in Machine Reading Comprehension}. In \bibinfo{booktitle}{\emph{Findings of the Association for Computational Linguistics: EMNLP 2020}}. \bibinfo{pages}{2496--2501}.
\newblock


\bibitem[Roemmele and Gordon(2015)]%
        {roemmele2015creative}
\bibfield{author}{\bibinfo{person}{Melissa Roemmele} {and} \bibinfo{person}{Andrew~S Gordon}.} \bibinfo{year}{2015}\natexlab{}.
\newblock \showarticletitle{Creative help: A story writing assistant}. In \bibinfo{booktitle}{\emph{Interactive Storytelling: 8th International Conference on Interactive Digital Storytelling, ICIDS 2015, Copenhagen, Denmark, November 30-December 4, 2015, Proceedings 8}}. Springer, \bibinfo{pages}{81--92}.
\newblock


\bibitem[Samuel et~al\mbox{.}(2016)]%
        {samuel2016design}
\bibfield{author}{\bibinfo{person}{Ben Samuel}, \bibinfo{person}{Michael Mateas}, {and} \bibinfo{person}{Noah Wardrip-Fruin}.} \bibinfo{year}{2016}\natexlab{}.
\newblock \showarticletitle{The design of Writing Buddy: a mixed-initiative approach towards computational story collaboration}. In \bibinfo{booktitle}{\emph{Interactive Storytelling: 9th International Conference on Interactive Digital Storytelling, ICIDS 2016, Los Angeles, CA, USA, November 15--18, 2016, Proceedings 9}}. Springer, \bibinfo{pages}{388--396}.
\newblock


\bibitem[Sharma(2021)]%
        {sharma2021generative}
\bibfield{author}{\bibinfo{person}{Sanjay Sharma}.} \bibinfo{year}{2021}\natexlab{}.
\newblock \showarticletitle{Generative Models for Question Generation}.
\newblock \bibinfo{journal}{\emph{Available at SSRN 3970263}} (\bibinfo{year}{2021}).
\newblock


\bibitem[Singh et~al\mbox{.}(2016)]%
        {singh2016expedition}
\bibfield{author}{\bibinfo{person}{Jaspreet Singh}, \bibinfo{person}{Wolfgang Nejdl}, {and} \bibinfo{person}{Avishek Anand}.} \bibinfo{year}{2016}\natexlab{}.
\newblock \showarticletitle{Expedition: a time-aware exploratory search system designed for scholars}. In \bibinfo{booktitle}{\emph{Proceedings of the 39th International ACM SIGIR conference on Research and Development in Information Retrieval}}. \bibinfo{pages}{1105--1108}.
\newblock


\bibitem[Sparck~Jones(1972)]%
        {sparck1972statistical}
\bibfield{author}{\bibinfo{person}{Karen Sparck~Jones}.} \bibinfo{year}{1972}\natexlab{}.
\newblock \showarticletitle{A statistical interpretation of term specificity and its application in retrieval}.
\newblock \bibinfo{journal}{\emph{Journal of documentation}} \bibinfo{volume}{28}, \bibinfo{number}{1} (\bibinfo{year}{1972}), \bibinfo{pages}{11--21}.
\newblock


\bibitem[Stiennon et~al\mbox{.}(2020)]%
        {stiennon2020learning}
\bibfield{author}{\bibinfo{person}{Nisan Stiennon}, \bibinfo{person}{Long Ouyang}, \bibinfo{person}{Jeffrey Wu}, \bibinfo{person}{Daniel Ziegler}, \bibinfo{person}{Ryan Lowe}, \bibinfo{person}{Chelsea Voss}, \bibinfo{person}{Alec Radford}, \bibinfo{person}{Dario Amodei}, {and} \bibinfo{person}{Paul~F Christiano}.} \bibinfo{year}{2020}\natexlab{}.
\newblock \showarticletitle{Learning to summarize with human feedback}.
\newblock \bibinfo{journal}{\emph{Advances in Neural Information Processing Systems}}  \bibinfo{volume}{33} (\bibinfo{year}{2020}), \bibinfo{pages}{3008--3021}.
\newblock


\bibitem[Sun et~al\mbox{.}(2024)]%
        {trustllm2024}
\bibfield{author}{\bibinfo{person}{Lichao Sun}, \bibinfo{person}{Yue Huang}, \bibinfo{person}{Haoran Wang}, \bibinfo{person}{Siyuan Wu}, \bibinfo{person}{Qihui Zhang}, \bibinfo{person}{Chujie Gao}, \bibinfo{person}{Yixin Huang}, \bibinfo{person}{Wenhan Lyu}, \bibinfo{person}{Yixuan Zhang}, \bibinfo{person}{Xiner Li}, \bibinfo{person}{Zhengliang Liu}, \bibinfo{person}{Yixin Liu}, \bibinfo{person}{Yijue Wang}, \bibinfo{person}{Zhikun Zhang}, \bibinfo{person}{Bhavya Kailkhura}, \bibinfo{person}{Caiming Xiong}, \bibinfo{person}{Chao Zhang}, \bibinfo{person}{Chaowei Xiao}, \bibinfo{person}{Chunyuan Li}, \bibinfo{person}{Eric Xing}, \bibinfo{person}{Furong Huang}, \bibinfo{person}{Hao Liu}, \bibinfo{person}{Heng Ji}, \bibinfo{person}{Hongyi Wang}, \bibinfo{person}{Huan Zhang}, \bibinfo{person}{Huaxiu Yao}, \bibinfo{person}{Manolis Kellis}, \bibinfo{person}{Marinka Zitnik}, \bibinfo{person}{Meng Jiang}, \bibinfo{person}{Mohit Bansal}, \bibinfo{person}{James Zou}, \bibinfo{person}{Jian Pei}, \bibinfo{person}{Jian
  Liu}, \bibinfo{person}{Jianfeng Gao}, \bibinfo{person}{Jiawei Han}, \bibinfo{person}{Jieyu Zhao}, \bibinfo{person}{Jiliang Tang}, \bibinfo{person}{Jindong Wang}, \bibinfo{person}{John Mitchell}, \bibinfo{person}{Kai Shu}, \bibinfo{person}{Kaidi Xu}, \bibinfo{person}{Kai-Wei Chang}, \bibinfo{person}{Lifang He}, \bibinfo{person}{Lifu Huang}, \bibinfo{person}{Michael Backes}, \bibinfo{person}{Neil~Zhenqiang Gong}, \bibinfo{person}{Philip~S. Yu}, \bibinfo{person}{Pin-Yu Chen}, \bibinfo{person}{Quanquan Gu}, \bibinfo{person}{Ran Xu}, \bibinfo{person}{Rex Ying}, \bibinfo{person}{Shuiwang Ji}, \bibinfo{person}{Tianlong Chen}, \bibinfo{person}{Tianming Liu}, \bibinfo{person}{Tianyi Zhou}, \bibinfo{person}{Willian Wang}, \bibinfo{person}{Xiang Li}, \bibinfo{person}{Xiangliang Zhang}, \bibinfo{person}{Xiao Wang}, \bibinfo{person}{Xing Xie}, \bibinfo{person}{Xun Chen}, \bibinfo{person}{Xuyu Wang}, \bibinfo{person}{Yan Liu}, \bibinfo{person}{Yanfang Ye}, \bibinfo{person}{Yinzhi Cao}, {and} \bibinfo{person}{Yue Zhao}.}
  \bibinfo{year}{2024}\natexlab{}.
\newblock \showarticletitle{TrustLLM: Trustworthiness in Large Language Models}. In \bibinfo{booktitle}{\emph{Proc. The Forty-first International Conference on Machine Learning (ICML2024)}}.
\newblock


\bibitem[Tam et~al\mbox{.}(2022)]%
        {tam2022evaluating}
\bibfield{author}{\bibinfo{person}{Derek Tam}, \bibinfo{person}{Anisha Mascarenhas}, \bibinfo{person}{Shiyue Zhang}, \bibinfo{person}{Sarah Kwan}, \bibinfo{person}{Mohit Bansal}, {and} \bibinfo{person}{Colin Raffel}.} \bibinfo{year}{2022}\natexlab{}.
\newblock \showarticletitle{Evaluating the Factual Consistency of Large Language Models Through Summarization}.
\newblock \bibinfo{journal}{\emph{arXiv preprint arXiv:2211.08412}} (\bibinfo{year}{2022}).
\newblock


\bibitem[Tian et~al\mbox{.}(2023)]%
        {tian2023fine}
\bibfield{author}{\bibinfo{person}{Katherine Tian}, \bibinfo{person}{Eric Mitchell}, \bibinfo{person}{Huaxiu Yao}, \bibinfo{person}{Christopher~D Manning}, {and} \bibinfo{person}{Chelsea Finn}.} \bibinfo{year}{2023}\natexlab{}.
\newblock \showarticletitle{Fine-tuning Language Models for Factuality}.
\newblock \bibinfo{journal}{\emph{arXiv preprint arXiv:2311.08401}} (\bibinfo{year}{2023}).
\newblock


\bibitem[Toniolo et~al\mbox{.}(2023)]%
        {toniolo2023human}
\bibfield{author}{\bibinfo{person}{Alice Toniolo}, \bibinfo{person}{Federico Cerutti}, \bibinfo{person}{Timothy~J Norman}, \bibinfo{person}{Nir Oren}, \bibinfo{person}{John~A Allen}, \bibinfo{person}{Mani Srivastava}, {and} \bibinfo{person}{Paul Sullivan}.} \bibinfo{year}{2023}\natexlab{}.
\newblock \showarticletitle{Human-machine collaboration in intelligence analysis: An expert evaluation}.
\newblock \bibinfo{journal}{\emph{Intelligent Systems with Applications}}  \bibinfo{volume}{17} (\bibinfo{year}{2023}), \bibinfo{pages}{200151}.
\newblock


\bibitem[Vaswani et~al\mbox{.}(2017)]%
        {vaswani2017attention}
\bibfield{author}{\bibinfo{person}{Ashish Vaswani}, \bibinfo{person}{Noam Shazeer}, \bibinfo{person}{Niki Parmar}, \bibinfo{person}{Jakob Uszkoreit}, \bibinfo{person}{Llion Jones}, \bibinfo{person}{Aidan~N Gomez}, \bibinfo{person}{{\L}ukasz Kaiser}, {and} \bibinfo{person}{Illia Polosukhin}.} \bibinfo{year}{2017}\natexlab{}.
\newblock \showarticletitle{Attention is all you need}.
\newblock \bibinfo{journal}{\emph{Advances in neural information processing systems}}  \bibinfo{volume}{30} (\bibinfo{year}{2017}).
\newblock


\bibitem[Wang et~al\mbox{.}(2022b)]%
        {wang2022schema}
\bibfield{author}{\bibinfo{person}{Hongwei Wang}, \bibinfo{person}{Zixuan Zhang}, \bibinfo{person}{Sha Li}, \bibinfo{person}{Jiawei Han}, \bibinfo{person}{Yizhou Sun}, \bibinfo{person}{Hanghang Tong}, \bibinfo{person}{Joseph~P Olive}, {and} \bibinfo{person}{Heng Ji}.} \bibinfo{year}{2022}\natexlab{b}.
\newblock \showarticletitle{Schema-guided event graph completion}.
\newblock \bibinfo{journal}{\emph{arXiv preprint arXiv:2206.02921}} (\bibinfo{year}{2022}).
\newblock


\bibitem[Wang et~al\mbox{.}(2019)]%
        {WangACL2019}
\bibfield{author}{\bibinfo{person}{Qingyun Wang}, \bibinfo{person}{Lifu Huang}, \bibinfo{person}{Zhiying Jiang}, \bibinfo{person}{Kevin Knight}, \bibinfo{person}{Heng Ji}, \bibinfo{person}{Mohit Bansal}, {and} \bibinfo{person}{Yi Luan}.} \bibinfo{year}{2019}\natexlab{}.
\newblock \showarticletitle{PaperRobot: Incremental Draft Generation of Scientific Ideas}. In \bibinfo{booktitle}{\emph{Proc. The 57th Annual Meeting of the Association for Computational Linguistics (ACL2019)}}.
\newblock


\bibitem[Wang et~al\mbox{.}(2020)]%
        {Wang2020}
\bibfield{author}{\bibinfo{person}{Qingyun Wang}, \bibinfo{person}{Qi Zeng}, \bibinfo{person}{Lifu Huang}, \bibinfo{person}{Kevin Knight}, \bibinfo{person}{Heng Ji}, {and} \bibinfo{person}{Nazneen~Fatema Rajani}.} \bibinfo{year}{2020}\natexlab{}.
\newblock \showarticletitle{ReviewRobot: Explainable Paper Review Generation based on Knowledge Synthesis}. In \bibinfo{booktitle}{\emph{Proc. The 13th International Conference on Natural Language Generation (INLG2020)}}.
\newblock


\bibitem[Wang et~al\mbox{.}(2023)]%
        {selfcollaboration2023}
\bibfield{author}{\bibinfo{person}{Zhenhailong Wang}, \bibinfo{person}{Shaoguang Mao}, \bibinfo{person}{Wenshan Wu}, \bibinfo{person}{Tao Ge}, \bibinfo{person}{Furu Wei}, {and} \bibinfo{person}{Heng Ji}.} \bibinfo{year}{2023}\natexlab{}.
\newblock \showarticletitle{Unleashing Cognitive Synergy in Large Language Models: A Task-Solving Agent through Multi-Persona Self-Collaboration}. In \bibinfo{booktitle}{\emph{arxiv}}.
\newblock


\bibitem[Wang et~al\mbox{.}(2022a)]%
        {wang2022towards}
\bibfield{author}{\bibinfo{person}{Zichao Wang}, \bibinfo{person}{Jakob Valdez}, \bibinfo{person}{Debshila Basu~Mallick}, {and} \bibinfo{person}{Richard~G Baraniuk}.} \bibinfo{year}{2022}\natexlab{a}.
\newblock \showarticletitle{Towards Human-Like Educational Question Generation with Large Language Models}. In \bibinfo{booktitle}{\emph{International Conference on Artificial Intelligence in Education}}. Springer, \bibinfo{pages}{153--166}.
\newblock


\bibitem[Wu et~al\mbox{.}(2019)]%
        {wu2019npa}
\bibfield{author}{\bibinfo{person}{Chuhan Wu}, \bibinfo{person}{Fangzhao Wu}, \bibinfo{person}{Mingxiao An}, \bibinfo{person}{Jianqiang Huang}, \bibinfo{person}{Yongfeng Huang}, {and} \bibinfo{person}{Xing Xie}.} \bibinfo{year}{2019}\natexlab{}.
\newblock \showarticletitle{NPA: neural news recommendation with personalized attention}. In \bibinfo{booktitle}{\emph{Proceedings of the 25th ACM SIGKDD international conference on knowledge discovery \& data mining}}. \bibinfo{pages}{2576--2584}.
\newblock


\bibitem[Yang et~al\mbox{.}(2015)]%
        {yang-etal-2015-wikiqa}
\bibfield{author}{\bibinfo{person}{Yi Yang}, \bibinfo{person}{Wen-tau Yih}, {and} \bibinfo{person}{Christopher Meek}.} \bibinfo{year}{2015}\natexlab{}.
\newblock \showarticletitle{{W}iki{QA}: A Challenge Dataset for Open-Domain Question Answering}. In \bibinfo{booktitle}{\emph{Proceedings of the 2015 Conference on Empirical Methods in Natural Language Processing}}. \bibinfo{publisher}{Association for Computational Linguistics}, \bibinfo{address}{Lisbon, Portugal}, \bibinfo{pages}{2013--2018}.
\newblock
\urldef\tempurl%
\url{https://doi.org/10.18653/v1/D15-1237}
\showDOI{\tempurl}


\bibitem[Young(2019)]%
        {overload1}
\bibfield{author}{\bibinfo{person}{Alex Young}.} \bibinfo{year}{2019}\natexlab{}.
\newblock \bibinfo{title}{Too Much Information: Ineffective Intelligence Collection}.
\newblock
\newblock
\newblock
\shownote{Harvard International Review. https://hir.harvard.edu/too-much-information/}.


\bibitem[Yuan et~al\mbox{.}(2022)]%
        {yuan2022wordcraft}
\bibfield{author}{\bibinfo{person}{Ann Yuan}, \bibinfo{person}{Andy Coenen}, \bibinfo{person}{Emily Reif}, {and} \bibinfo{person}{Daphne Ippolito}.} \bibinfo{year}{2022}\natexlab{}.
\newblock \showarticletitle{Wordcraft: story writing with large language models}. In \bibinfo{booktitle}{\emph{27th International Conference on Intelligent User Interfaces}}. \bibinfo{pages}{841--852}.
\newblock


\bibitem[Zhang et~al\mbox{.}(2020)]%
        {zhang2020pegasus}
\bibfield{author}{\bibinfo{person}{Jingqing Zhang}, \bibinfo{person}{Yao Zhao}, \bibinfo{person}{Mohammad Saleh}, {and} \bibinfo{person}{Peter Liu}.} \bibinfo{year}{2020}\natexlab{}.
\newblock \showarticletitle{Pegasus: Pre-training with extracted gap-sentences for abstractive summarization}. In \bibinfo{booktitle}{\emph{International Conference on Machine Learning}}. PMLR, \bibinfo{pages}{11328--11339}.
\newblock


\bibitem[Zhang et~al\mbox{.}(2023)]%
        {zhang2023benchmarking}
\bibfield{author}{\bibinfo{person}{Tianyi Zhang}, \bibinfo{person}{Faisal Ladhak}, \bibinfo{person}{Esin Durmus}, \bibinfo{person}{Percy Liang}, \bibinfo{person}{Kathleen McKeown}, {and} \bibinfo{person}{Tatsunori~B Hashimoto}.} \bibinfo{year}{2023}\natexlab{}.
\newblock \showarticletitle{Benchmarking Large Language Models for News Summarization}.
\newblock \bibinfo{journal}{\emph{arXiv preprint arXiv:2301.13848}} (\bibinfo{year}{2023}).
\newblock


\bibitem[Zhang et~al\mbox{.}(2021)]%
        {zhang2021joint}
\bibfield{author}{\bibinfo{person}{Zeyu Zhang}, \bibinfo{person}{Thuy Vu}, {and} \bibinfo{person}{Alessandro Moschitti}.} \bibinfo{year}{2021}\natexlab{}.
\newblock \showarticletitle{Joint Models for Answer Verification in Question Answering Systems}. In \bibinfo{booktitle}{\emph{Proceedings of the 59th Annual Meeting of the Association for Computational Linguistics and the 11th International Joint Conference on Natural Language Processing (Volume 1: Long Papers)}}. \bibinfo{pages}{3252--3262}.
\newblock


\end{thebibliography}
